\documentclass[prl,floatfix,twoside,showpacs,byrevtex,superscriptaddress]{revtex4-1}

\usepackage{currfile}

\usepackage{graphicx,epsfig,verbatim,enumerate}
\usepackage{amssymb,amsmath}
\usepackage{ifthen}
\newboolean{twocolswitch}

\newcommand{\sindex}[1]{}
\newcommand{\nindex}[1]{}

\newcommand{\www}[1]{\url{#1}}

\usepackage{lettrine}

\usepackage{color}

\usepackage{amsmath}
\usepackage{amssymb}
\usepackage{bigints}
\usepackage{graphicx}
\usepackage{caption}
\usepackage{float}

\usepackage{textpos}

\usepackage[table]{xcolor}

\setboolean{twocolswitch}{false}

\begin{document}

\title{
  A Sentiment Analysis of Breast Cancer Treatment Experiences and Healthcare Perceptions Across Twitter
}

\author{
\firstname{Eric M.}
\surname{Clark}
}

\email{eclark@uvm.edu}

\affiliation{Department of Mathematics and \& Statistics,
  The University of Vermont,
  Burlington,VT, USA. }

\affiliation{Vermont Complex Systems Center,
  The University of Vermont,
  Burlington,VT, USA .}

\affiliation{Vermont Advanced Computing Core,
  The University of Vermont,
  Burlington,VT, USA .}
  
\affiliation{Computational Story Lab,
  The University of Vermont,
  Burlington,VT, USA .}
  
  \affiliation{SEGS Lab,
  The University of Vermont,
  Burlington,VT, USA .}

 \author{
\firstname{Ted}
\surname{James}
}

  \affiliation{Breast Care Center,
  Harvard Medical Center,
  Boston,MA, USA .}

   \author{
\firstname{Chris A.}
\surname{Jones}
}
  
  \affiliation{UVM Health Network,
  The University of Vermont,
  Burlington,VT, USA .}
  
 \affiliation{Arizona State University, Phoenix, AZ}

     \author{
\firstname{Amulya}
\surname{Alapati}
}

  \affiliation{Department of Surgery, 
  Beth Israel Deaconess Medical Center, 
  Boston , MA, USA}

   \author{
\firstname{Promise}
\surname{Ukandu}
}

     \affiliation{Department of Orthopedics, 
  Beth Israel Deaconess Medical Center, 
  Boston , MA, USA}

   \author{
\firstname{Christopher M.}
\surname{Danforth}
}
\affiliation{Department of Mathematics and \& Statistics,
  The University of Vermont,
  Burlington,VT, USA. }

\affiliation{Vermont Complex Systems Center,
  The University of Vermont,
  Burlington,VT, USA .}

\affiliation{Vermont Advanced Computing Core,
  The University of Vermont,
  Burlington,VT, USA .}
  
\affiliation{Computational Story Lab,
  The University of Vermont,
  Burlington,VT, USA .}

   \author{
\firstname{Peter Sheridan}
\surname{Dodds}
}
\affiliation{Department of Mathematics and \& Statistics,
  The University of Vermont,
  Burlington,VT, USA. }

\affiliation{Vermont Complex Systems Center,
  The University of Vermont,
  Burlington,VT, USA .}

\affiliation{Vermont Advanced Computing Core,
  The University of Vermont,
  Burlington,VT, USA .}
  
\affiliation{Computational Story Lab,
  The University of Vermont,
  Burlington,VT, USA .}

\date{\today}

\begin{abstract}
  \vspace{1mm}
\hspace{-5mm} \textbf{Background:} Social media has the capacity to afford the healthcare industry with valuable feedback from patients who reveal and express their medical decision-making process, as well as self-reported quality of life indicators both during and post treatment. In prior work, \citet{crannell2016pattern}, we have studied an active cancer patient population on Twitter and compiled a set of tweets describing their experience with this disease. We refer to these online public testimonies as ``Invisible Patient Reported Outcomes'' (iPROs), because they carry relevant indicators, yet are difficult to capture by conventional means of self-report. \\
\textbf{Methods:} Our present study aims to identify tweets related to the patient experience as an additional informative tool for monitoring public health.  Using Twitter's public streaming API, we compiled over 5.3 million ``breast cancer'' related tweets spanning September 2016 until mid December 2017.  We combined supervised machine learning methods with natural language processing to sift tweets relevant to breast cancer patient experiences.  We analyzed a sample of 845 breast cancer patient and survivor accounts, responsible for over 48,000 posts. We investigated tweet content with a hedonometric sentiment analysis to quantitatively extract emotionally charged topics.  \\
\textbf{Results:} We found that positive experiences were shared regarding patient treatment, raising support, and spreading awareness.  Further discussions related to healthcare were prevalent and largely negative focusing on fear of political legislation that could result in loss of coverage.  \\
\textbf{Conclusions:} Social media can provide a positive outlet for patients to discuss their needs and concerns regarding their healthcare coverage and treatment needs.  Capturing iPROs from online communication can help inform healthcare professionals and lead to more connected and personalized treatment regimens.  
  
\end{abstract}

\maketitle

\section{Introduction}

  Twitter has shown potential for monitoring public health trends, \citet{alajajian2017lexicocalorimeter,paul2011you,shive2013perspectives,dredze2012social,reece2016forecasting}, disease surveillance, \citet{lamb2013separating}, and providing a rich online forum for cancer patients, \citet{sugawara2012cancer}. Social media has been validated as an effective educational and support tool for breast cancer patients, \citet{attai2015twitter}, as well as for generating awareness, \citet{bender2011seeking}. Successful supportive organizations use social media sites for patient interaction, public education, and donor outreach, \citet{fussell2013communicating}. The advantages, limitations, and future potential of using social media in healthcare has been thoroughly reviewed, \citet{moorhead2013new}. Our study aims to investigate tweets mentioning ``breast'' and ``cancer" to analyze patient populations and selectively obtain content relevant to patient treatment experiences. 

Our previous study, \citet{crannell2016pattern}, collected tweets mentioning ``cancer'' over several months to investigate the potential for monitoring self-reported patient treatment experiences. Non-relevant tweets (e.g. astrological and horoscope references) were removed and the study identified a sample of 660 tweets from patients who were describing their condition. These self-reported diagnostic indicators allowed for a sentiment analysis of tweets authored by patients. However, this process was tedious, since the samples were hand verified and sifted through multiple keyword searches.  Here, we aim to automate this process with machine learning context classifiers in order to build larger sets of patient self-reported outcomes in order to quantify the patent experience. 

Patients with breast cancer represent a majority of people affected by and living with cancer. As such, it becomes increasingly important to learn from their experiences and understand their journey from their own perspective. The collection and analysis of invisible patient reported outcomes (iPROs) offers a unique opportunity to better understand the patient perspective of care and identify gaps meeting particular patient care needs. 

\section{Methods}

\subsection{Data Description}

\hspace{5mm} Twitter provides a free streaming Application Programming Interface (API), \cite{TwitterStreaming}, for researchers and developers to mine samples of public tweets.  Language processing and data mining, \citet{roesslein2009tweepy}, was conducted using the Python programming language.  The free public API allows targeted keyword mining of up to 1\% of Twitter's full volume at any given time, referred to as the `Spritzer Feed'.

\begin{figure}[H]
  \hskip-1.32cm
  \includegraphics[scale=.25]{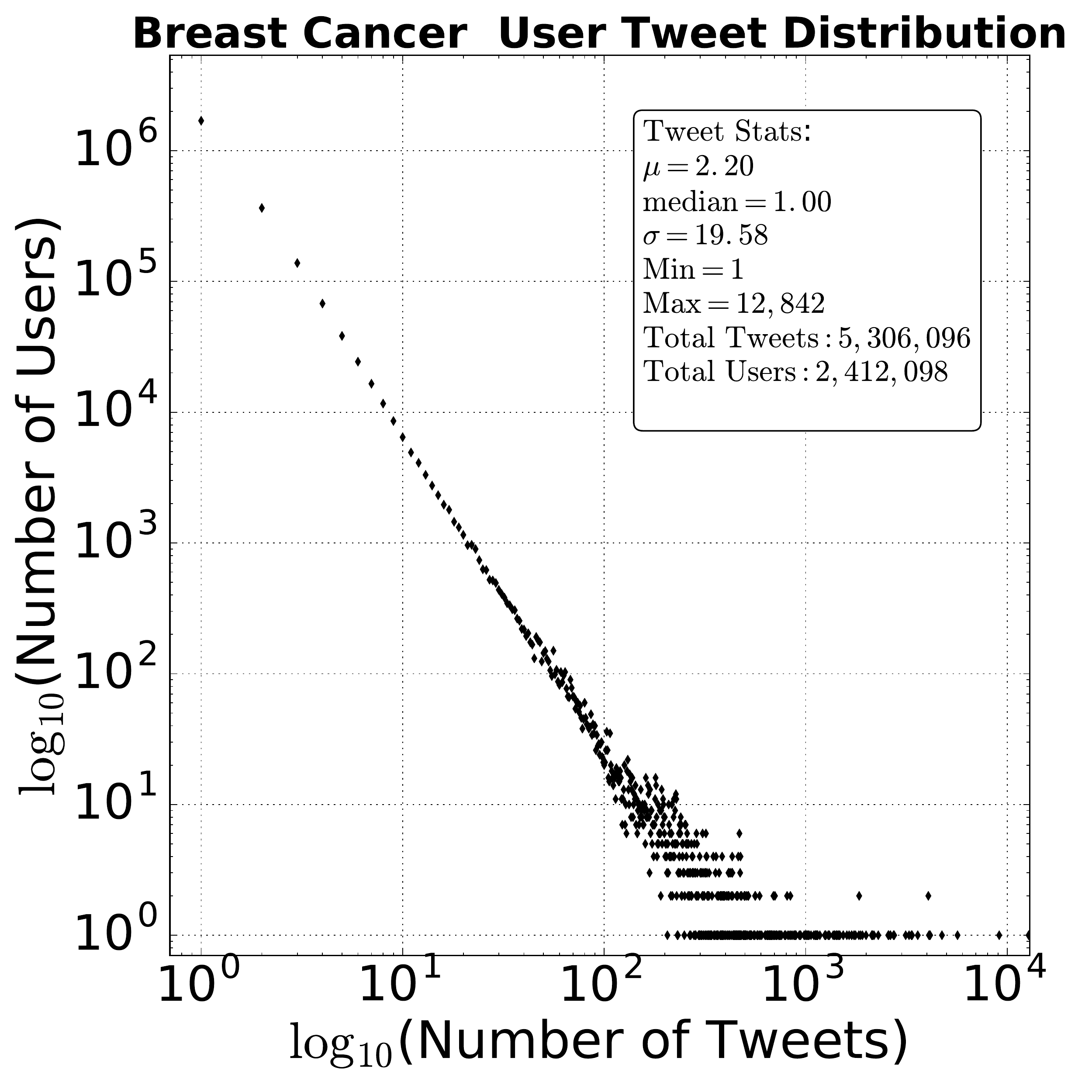}
 \includegraphics[scale=.32]{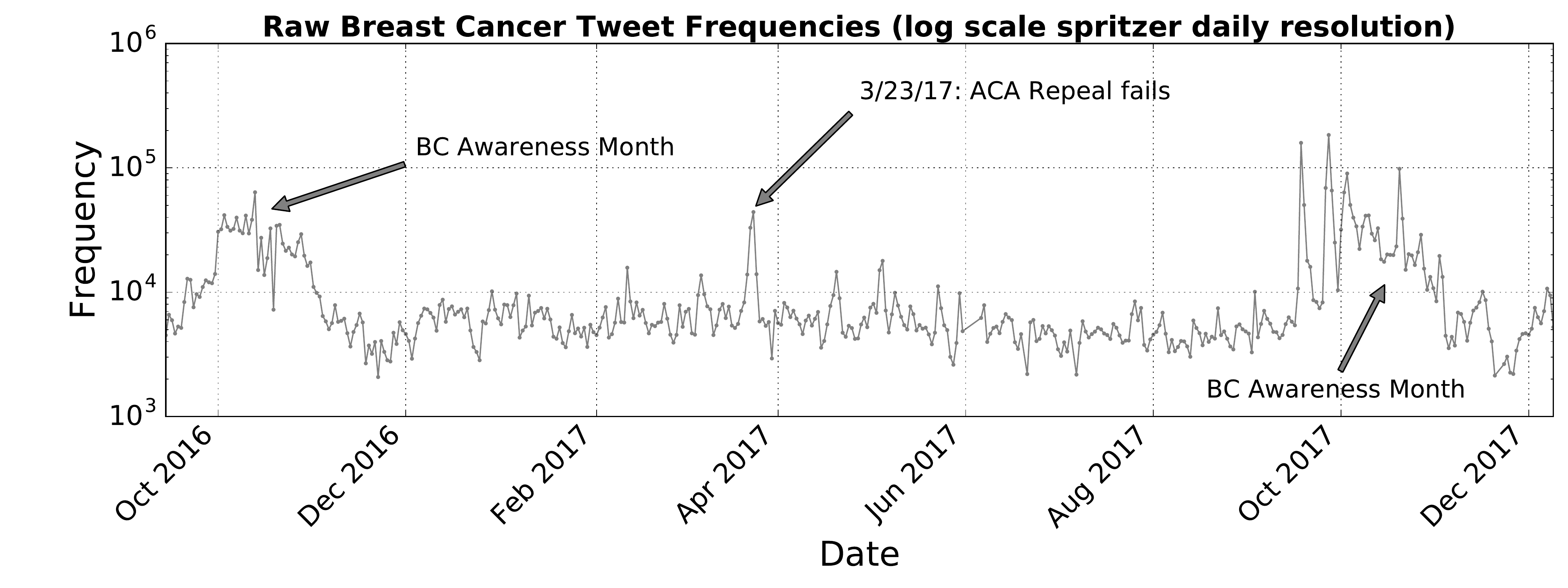}
 \caption{ (left) The distribution of tweets per given user is plotted on a log axis. The tail tends to be high frequency automated accounts, some of which provide daily updates or news related to cancer.   (right) A frequency time-series of the tweets collected, binned by day. }
\label{fig:TweetFeatures}
 \end{figure}

 \hspace{5mm} We collected tweets from two distinct Spritzer endpoints from September 15th, 2016 through December 9th, 2017.  The primary feed for the analysis collected $5.3$ million tweets containing the keywords `breast'  AND `cancer'.  See Figure \ref{fig:TweetFeatures} for detailed Twitter frequency statistics along with the user activity distribution.  Our secondary feed searched just for the keyword `cancer' which served as a comparison ( $76.4$ million tweets,  see Appendix 1), and helped us collect additional tweets relevant to cancer from patients.  The numeric account ID provided in tweets helps to distinguish high frequency tweeting entities.  

Sentence classification combines natural language processing (NLP) with machine learning to identify trends in sentence structure, \citet{zhang2015sensitivity,blunsom2014convolutional}. Each tweet is converted to a numeric word vector in order to identify distinguishing features by training an NLP classifier on a validated set of relevant tweets. The classifier acts as a tool to sift through ads, news, and comments not related to patients. Our scheme combines a logistic regression classifier, \citet{genkin2007large}, with a Convolutional Neural Network (CNN), \citet{kim2014convolutional,britz2015implementing}, to identify self-reported diagnostic tweets. 

It is important to be wary of automated accounts (e.g. bots, spam) whose large output of tweets pollute relevant organic content, \citet{clark2016sifting}, and can distort sentiment analyses, \citet{clark2016vaporous}. Prior to applying sentence classification, we removed tweets containing hyperlinks to remove automated content (some organic content is necessarily lost with this strict constraint). 

 The user tweet distribution in Figure  \ref{fig:TweetFeatures}, shows the number of users as a function of the number of their tweets we collected.  With an average frequency of $2.2$ tweets per user, this is a relatively healthy activity distribution.  High frequency tweeting accounts are present in the tail, with a single account producing over 12,000 tweets \textemdash an automated account served as a support tool called `ClearScan' for patients in recovery. Approximately  98\% of the 2.4 million users shared less than 10 posts,  which accounted for 70\% of all sampled tweets. 

  The Twitter API also provided the number of tweets withheld from our sample, due to rate limiting.  Using these overflow statistics, we estimated the sampled proportion of tweets mentioning these keywords.  These targeted feeds were able to collect a large sample of all tweets mentioning these terms; approximately 96\% of tweets mentioning ``breast,cancer'' and 65.2\% of all tweets mentioning `cancer' while active.  More information regarding the types of Twitter endpoints and calculating the sampling proportion of collected tweets is described in Appendix II. 

  Our goal was to analyze content authored only by patients.  To help ensure this outcome we removed posts containing a URL for classification, \citet{clark2016sifting}.  Twitter allows users to spread content from other users via `retweets'.  We also removed these posts prior to classification to isolate tweets authored by patients.  We also accounted for non-relevant astrological content by removing all tweets containing any of the following horoscope indicators: `astrology',`zodiac',`astronomy',`horoscope',`aquarius',`pisces',`aries',`taurus',`leo',`virgo',`libra', and `scorpio'.  We preprocessed tweets by lowercasing and removing punctuation.  We also only analyzed tweets for which Twitter had identified `en' for the language English.    
  
  \vspace{6cm}

\pagebreak

\subsection{Sentiment Analysis and Hedonometrics}

 \hspace{5mm} We evaluated tweet sentiments with hedonometrics, \citet{dodds2011temporal,dodds2014human}, using LabMT, a labeled set of 10,000 frequently occurring words rated on a `happiness' scale by individuals  contracted through Amazon Mechanical Turk, a crowd-sourced  survey tool.   These happiness scores helped quantify the average emotional rating of text by totaling the scores from applicable words and normalizing by their total frequency.  Hence, the average happiness score, $h_{\textnormal{avg}}$, of a corpus with $N$ words in common with LabMT was computed with the weighted arithmetic mean of each word's frequency, $f_w$, and associated happiness score, $h_w$:  
 
\begin{equation} h_{\textnormal{avg}} =  \dfrac{ \sum \limits_{w=1}^{N} f_w \cdot h_w}{\sum \limits_{w=1}^N f_w}   \end{equation}

   The average happiness of each word was rated on a 9 point scale ranging from extremely negative (e.g., `emergency' 3.06, `hate' 2.34, `die' 1.74) to positive (e.g., `laughter' 8.50, `love' 8.42, `healthy' 8.02).
Neutral `stop words' ($4\leq h_{\textnormal{avg}}\leq 6$, e.g., `of','the', etc.) were removed to enhance the emotional signal of each set of tweets.  These high frequency, low sentiment words can dampen a signal, so their removal can help identify hidden trends.  One application is to plot $h_{\textnormal{avg}}$ as a function of time.  The happiness time-series can provide insight driving emotional content in text.  In particular, peak and dips (i.e., large deviations from the average) can help identify interesting themes that may be overlooked in the frequency distribution.   Calculated scores can give us comparative insight into the context between sets of tweets.

  ``Word shift graphs'' introduced in, \citet{dodds2011temporal},  compare the terms contributing to shifts in a computed word happiness from two term frequency distributions.  This tool is useful in isolating emotional themes from large sets of text and has been previously validated in monitoring public opinion, \citet{cody2015climate} as well as for geographical sentiment comparative analyses, \citet{mitchell2013geography}. See Appendix III for a general description of word shift graphs and how to interpret them.  

\subsection{ Relevance Classification: Logistic Model and CNN Architecture}

  We began by building a validated training set of tweets for our sentence classifier.  We compiled the patient tweets verified by, \citet{crannell2016pattern}, to train a logistic regression content relevance classifier using a similar framework as, \citet{genkin2007large}.  To test the classifier, we  compiled over 5 million tweets mentioning the word cancer from a  10\% `Gardenhose' random sample of Twitter spanning January through December 2015.   See Appendix 1 for a statistical overview of this corpus. 
 
 We tested a maximum entropy logistic regression classifier using a similar scheme as, \citet{genkin2007large}.  NLP classifiers operate by converting sentences to word vectors for identifying key characteristics --- the vocabulary of the classifier.  Within the vocabulary, weights were assigned to each word based upon a frequency statistic.  We used the term frequency crossed with the inverse document frequency (tf-idf), as described in , \citet{genkin2007large}.   The tf-idf weights helped distinguish each term's relative weight across the entire corpus,  instead of relying on raw frequency. This statistic dampens highly frequent non-relevant words (e.g. `of', `the', etc.) and enhances relatively rare yet informative terms (e.g. survivor, diagnosed, fighting).  This method is commonly implemented in information retrieval for text mining, \citet{salton1975vector}.  The logistic regression context classifier then performs a binary classification of the tweets we collected from 2015.  See Appendix IV for an expanded description of the sentence classification methodology.  
\begin{table}[H]
\hskip3cm
\begin{tabular}{|c|l|}
\cline{1-2}
No. & Tweet Key Identifying Phrases \\ 
\hline
\hline

1 & Breast cancer fear gone! Tumor removed ... \\
\hline
2 & ... my tremendously difficult journey through Stage IV  Breast Cancer ... \\
\hline
3 &  ... life after breast cancer. I am 11 years Cancer Free ...  \\
\hline

4 & ... IM FIGHTING BREAST CANCER STAGE 3 ... \\

\hline
5 & @USER just got diagnosed with breast cancer ...\\

\hline
\hline
\end{tabular}
\caption{ {\textbf{Diagnostic Training Sample Tweet Phrases:}  A sample of self-reported diagnostic phrases from tweets used to train the logistic regression content classifier (modified to preserve anonymity).}} 
\end{table}

   We validated the logistic model's performance by manually verifying 1,000 tweets that were classified as `relevant'.  We uncovered three categories of immediate interest including: tweets authored by patients regarding their condition (21.6\%), tweets from friends/family with a direct connection to a patient (21.9\%),  and survivors in remission (8.8\%).  We also found users posting diagnostic related inquiries (7.6\%) about  possible symptoms that could be linked to breast cancer, or were interested in receiving preventative check-ups.  The rest (40.2\%) were related to `cancer', but not to patients and include public service updates as well as non-patient authored content (e.g., support groups).  We note that the classifier was trained on very limited validated data (N=660), which certainly impacted the results.  We used this validated annotated set of tweets to train a more sophisticated classifier to uncover self-diagnostic tweets from users describing their personal breast cancer experiences as current patients or survivors.

We implemented the Convolutional Neural Network (CNN) with Google's Tensorflow interface, \citet{abadi2016tensorflow}.  
We adapted our framework from, \citet{britz2015implementing}, but instead trained the CNN on these 1000 labeled cancer related tweets.  The trained CNN was applied to predict patient self-diagnostic tweets from our breast cancer dataset.  The CNN outputs a binary value: positive for a predicted tweet relevant to patients or survivors and negative for these other described categories (patient connected, unrelated, diagnostic inquiry).   The Tensorflow CNN interface reported a  $97.6\%$ accuracy when evaluating this set of labels with our trained model.  These labels were used to predict self-reported diagnostic tweets relevant to breast cancer patients.

 \section{Results}

     \hspace{5mm} A set of 845 breast cancer patient self-diagnostic Twitter profiles was compiled by implementing our logistic model followed by prediction with the trained CNN on 9 months of tweets.  The logistic model sifted 4,836 relevant tweets of which 1,331 were predicted to be self-diagnostic by the CNN.  Two independent groups annotated the 1,331 tweets to identify patients and evaluate the classifier's results.  The raters, showing high inter-rater reliability,  individually evaluated each tweet as self-diagnostic of a breast cancer patient or survivor. The rater's independent annotations had a 96\% agreement.
  
\begin{figure}[H]
  \hskip-1.32cm
  \includegraphics[scale=.25]{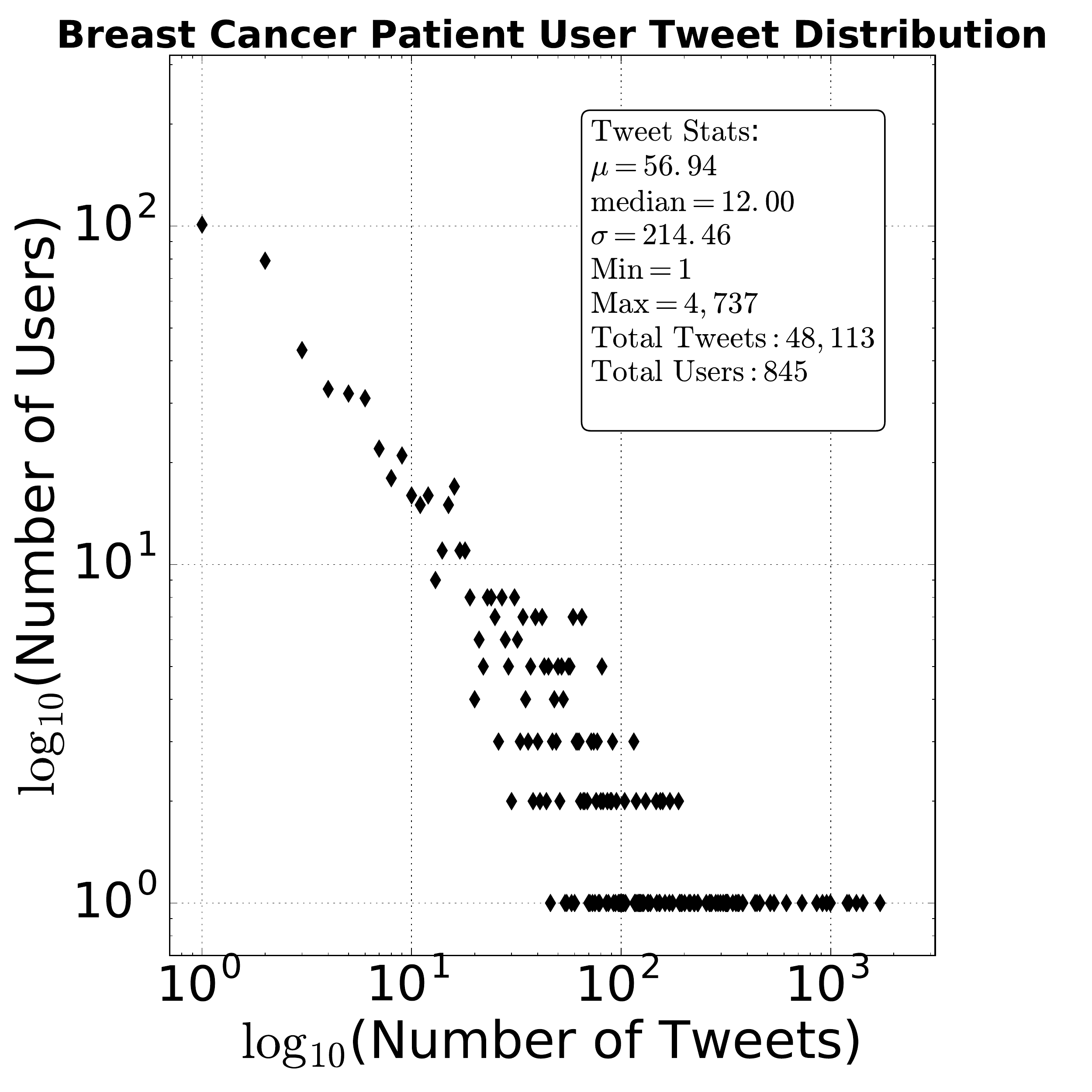}
 \includegraphics[scale=.32]{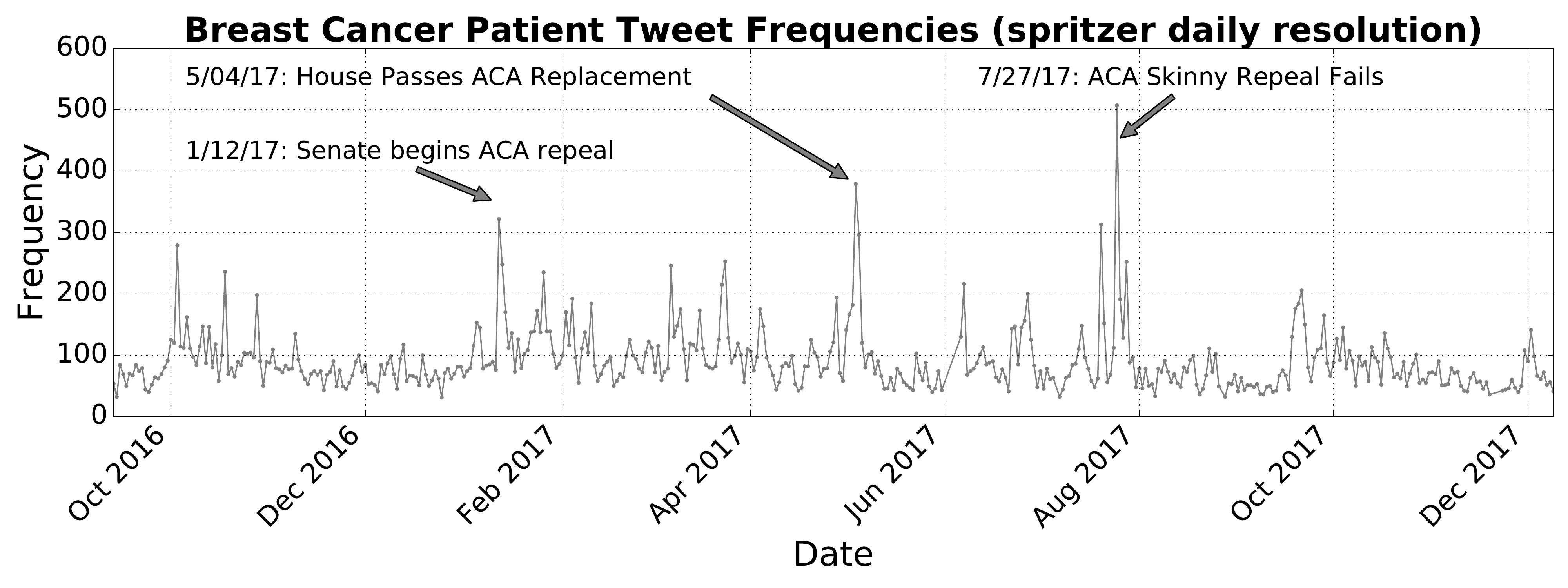}
 \caption{ (left) The distribution of tweets per given patient/survivor is plotted on a log axis along with a statistical summary of patient tweeting behavior. (right) A frequency time-series of patient tweets collected, binned by day.  }
\label{fig:PatientTweetFeatures}
 \end{figure}

             \hspace{5mm} The classifier correctly identified 1,140 tweets (85.6\%) from 845 profiles.  A total of 48,113 tweets from these accounts were compiled from both the `cancer'  (69\%) and `breast' `cancer' (31\%) feeds.  We provided tweet frequency statistics in Figure \ref{fig:PatientTweetFeatures}.  This is an indicator that this population of breast cancer patients and survivors are actively tweeting about topics related to `cancer' including their experiences and complications. 

\begin{figure}[H]
      \includegraphics[scale=.48]{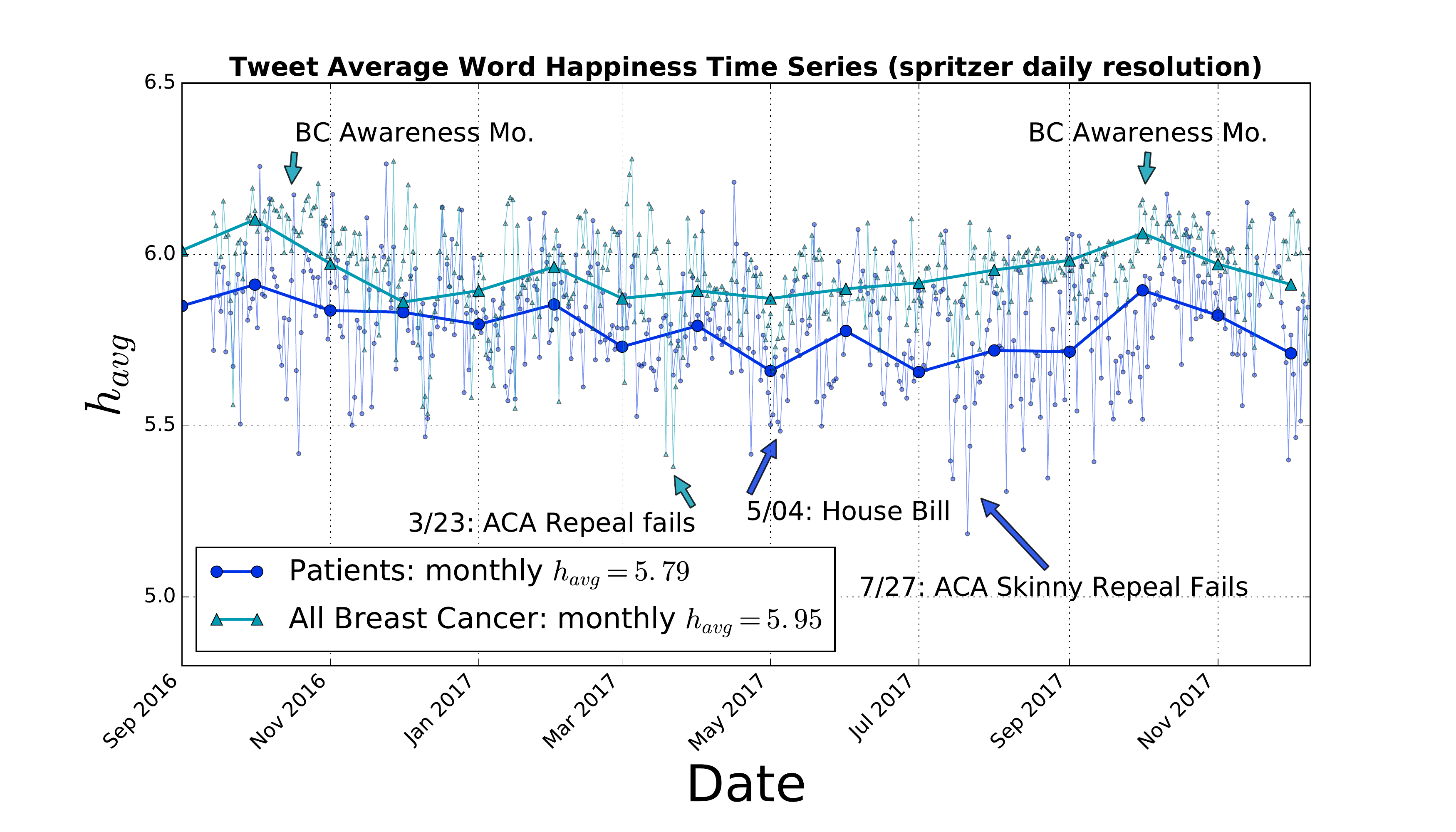}
   \caption{  Computed average word happiness as a function of day (small markers) and month (large markers) for both the `breast',`cancer'  and patient distributions.  The patient monthly average was less positive than the reference distribution ($h_{\textnormal{avg}} = 5.78$ v. $5.93$).
 }
\label{fig:TweetSentiments}
 \end{figure}

Next, we applied hedonometrics to compare the patient posts with all collected breast cancer tweets.  We found that the surveyed patient tweets were less positive than breast cancer reference tweets.   In Figure \ref{fig:TweetSentiments},  the time series plots computed average word happiness at monthly and daily resolutions.  The daily happiness scores (small markers) have a high fluctuation, especially within the smaller patient sample (average 100 tweets/day) compared to the reference distribution (average 10,000 tweets/day).  The monthly calculations (larger markers) highlight the negative shift in  average word happiness between the patients and reference tweets. Large fluctuations in computed word happiness correspond to noteworthy events, including breast cancer awareness month in October, cancer awareness month in February, as well as political debate regarding healthcare beginning in March May and July 2017.

\begin{figure}[H]
     \includegraphics[scale=.46]{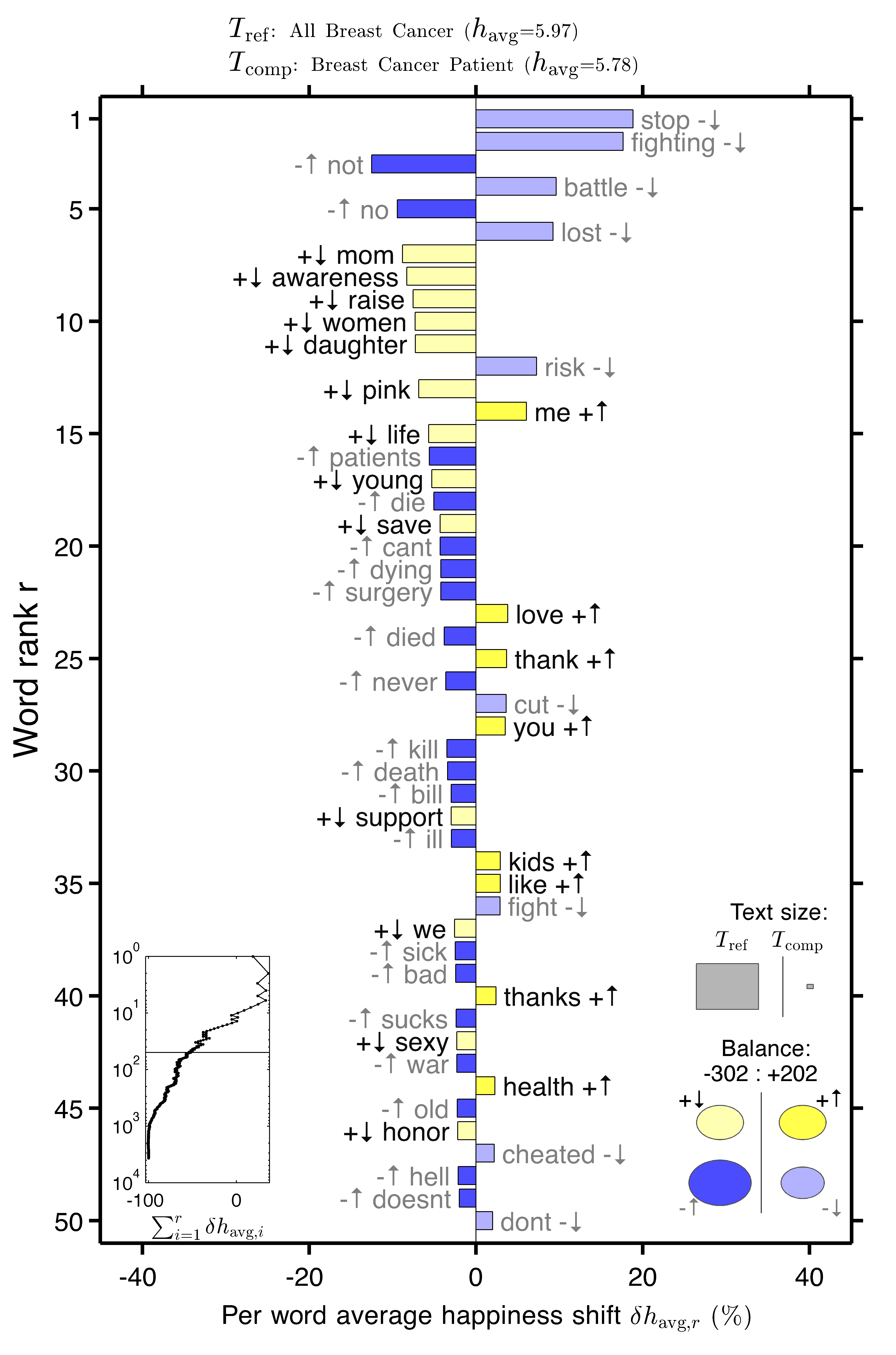}
  \includegraphics[scale=.46]{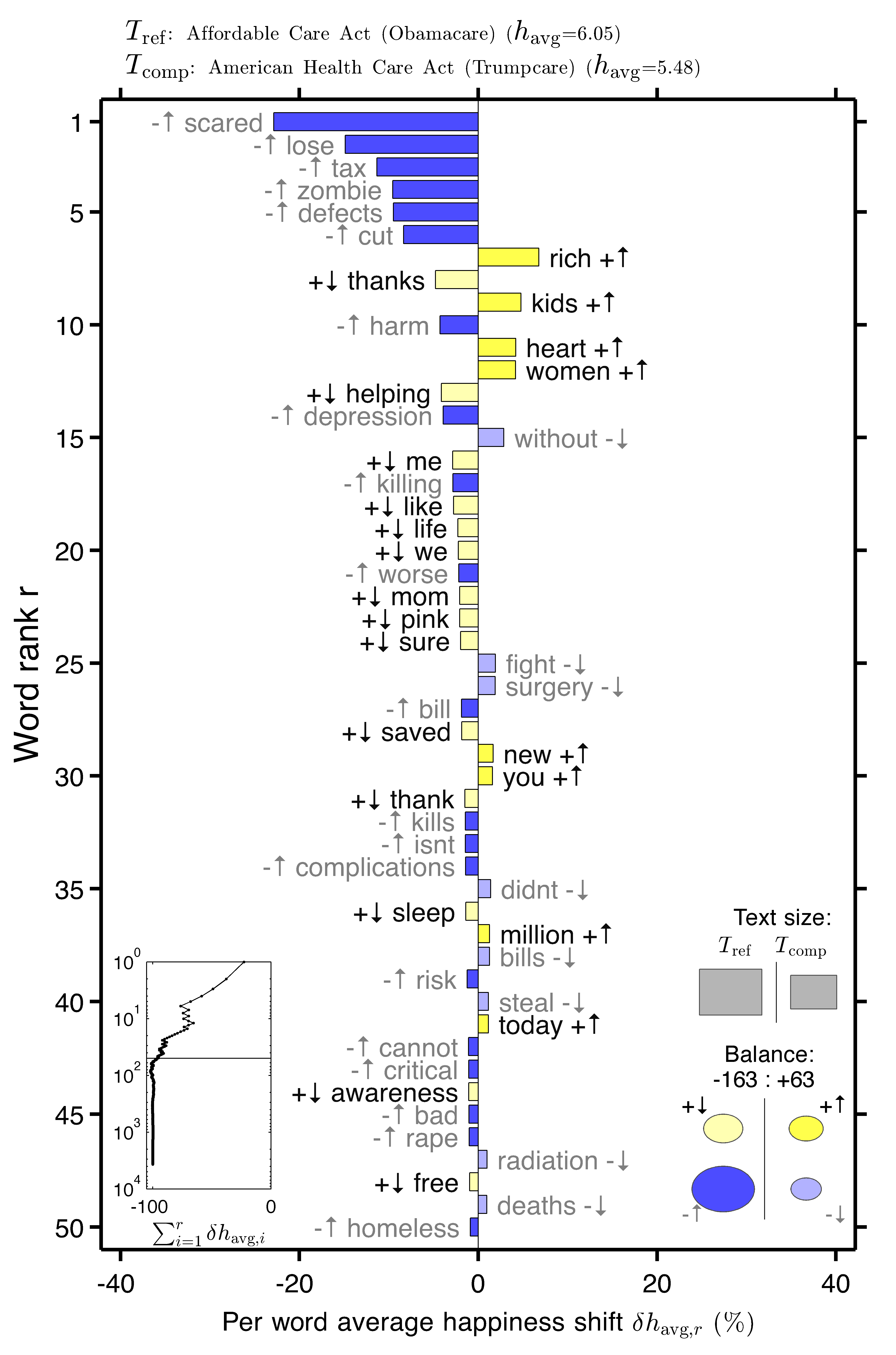}
 \caption{ (Left) Word shift graph comparing collected Breast Cancer Patient Tweets, $T_{\textnormal{comp}}$, to all Breast Cancer Tweets, $T_{\textnormal{ref}}$.  Patient Tweets were less positive ($h_{\textnormal{avg}} = 5.78$ v. $5.97$), due to a decrease in positive words `mom', `raise', `awareness', `women', `daughter', `pink', and `life' as well as  an increase in the negative words `no(t)', `patients, `dying', `killing', `surgery' `sick', `sucks',  and `bill'. (Right) Word shift graph comparing tweets mentioning the American Healthcare Act (AHCA, 10.5k tweets) to the Affordable Care Act (ACA, 16.9k tweets).  AHCA tweets were more negative ($h_{\textnormal{avg}} = 5.48$ v. $6.05$) due to a relative increase in the negative words `scared', `lose', `zombie', `defects', `depression', `harm', `killing', and `worse'.
 }
\label{fig:TweetWordshifts}
 \end{figure}

In Figure \ref{fig:TweetWordshifts} word shift  graphs display the top 50 words responsible for the shift in computed word happiness between distributions.  On the left, tweets from patients were compared to all collected breast cancer tweets.  Patient tweets, $T_{\textnormal{comp}}$, were less positive  ($h_{\textnormal{avg}} = 5.78$ v.  $5.97$) than the reference distribution,$ T_{\textnormal{ref}}$.  There were relatively less positive words `mom', `raise', `awareness', `women', `daughter', `pink', and `life' as well as  an increase in the negative words `no(t)', `patients, `dying', `killing', `surgery' `sick', `sucks', and `bill'.  Breast cancer awareness month, occurring in October, tends to be a high frequency period with generally more positive and supportive tweets from the general public which may account for some of the negative shift.  Notably, there was a relative increase of the positive words `me', `thank', `you' ,'love', and `like' which may indicate that many tweet contexts were from the patient's perspective regarding positive experiences. Many tweets regarding treatment were enthusiastic, supportive, and proactive.  Other posts were descriptive: over 165 sampled patient tweets mentioned personal chemo therapy experiences and details regarding their treatment schedule, and side effects.
  
  \hspace{5mm} Numerous patients and survivors in our sample had identified their condition in reference to the American healthcare regulation debate.    Many sampled views of the proposed legislation were very negative, since repealing the Affordable Care Act without replacement could leave many uninsured.   Other tweets mentioned worries regarding insurance premiums and costs for patients and survivors' continued screening.  In particular the pre-existing condition mandate was a chief concern of patients/survivors future coverage. This was echoed  by 55 of the sampled patients with the hashtag \#iamapreexistingcondition (See Table \ref{table:Hashtags}).
 
  Hashtags (\#)  are terms that categorize topics within posts.  In Table \ref{table:Hashtags}, the most frequently occurring hashtags from both the sampled patients (right) and full breast cancer corpus (left).  Each entry contains the tweet frequency, number of distinct profiles, and the relative happiness score ($h_{\textnormal{avg}}$) for comparisons. Political terms were prevalent in both distributions describing the Affordable Care Act (\#aca, \#obamacare, \#saveaca, \#pretectourcare) and the newly introduced American Healthcare Act (\#ahca, \#trumpcare). A visual representation of these hashtags are displayed using a word-cloud in the Appendix (Figure A4).
  
   Tweets referencing the AHCA were markedly more negative than those referencing the ACA.  This shift was investigated in Figure \ref{fig:TweetWordshifts} with a word shift graph.   We compared American Healthcare Act Tweets, $T_{\textnormal{comp}}$, to posts mentioning the Affordable Care Act, $T_{\textnormal{ref}}$.  AHCA were relatively more negative ($h_{\textnormal{avg}} = 5.48$ v. $6.05$) due to an increase of negatively charged words `scared', `lose', `tax', `zombie', `defects', `cut', `depression', `killing', and `worse' .  These were references to the bill leaving many patients/survivors without insurance and jeopardizing future treatment options.  `Zombie' referenced the bill's potential return for subsequent votes.
     
  \pagebreak
  \begin{table}[H]
\begin{tabular}{|c|c|c|c|c|}
\cline{1-5}
\multicolumn{5}{|c|}{  \normalsize{\textbf{Top Hashtags(\#): All Breast Cancer}}}\\ \hline \hline
Rank & Term & Tweets  & Users & $h_{\textnormal{avg}}$ \\ 
\hline 
1 & \#cancer & 67,111 & 23,171 & \textcolor{blue}{5.92} \\ 
2 & \#breastcancer & 66,400 & 22,247 & \textcolor{orange}{5.97} \\ 
3 & \#breast & 35,544 & 11,115 & \textcolor{orange}{6.0} \\ 
4 & \#nobraday & 23,406 & 16,785 & \textcolor{blue}{5.76} \\ 
5 & \#breastcancerawarenessmonth & 20,961 & 13,491 & \textcolor{orange}{6.06} \\ 
6 & \#health & 17,484 & 5,696 & \textcolor{blue}{5.82} \\ 
7 & \#twibbon & 16,809 & 14,332 & \textcolor{orange}{6.18} \\ 
8 & \#bcsm & 14,955 & 4,644 & \textcolor{blue}{5.95} \\ 
9 & \#survivor & 14,500 & 1,107 & \textcolor{orange}{5.98} \\ 
10 & \#idrivefor & 13,562 & 8,331 & \textcolor{orange}{6.06} \\ 
11 & \#breastcancerawareness & 13,429 & 8,820 & \textcolor{orange}{6.13} \\ 
12 & \#lymphedema & 13,263 & 2,274 & \textcolor{blue}{5.88} \\ 
13 & \#walk & 9,344 & 246 & \textcolor{orange}{6.0} \\ 
14 & \#aca & 8,903 & 8,105 & \textcolor{orange}{6.05} \\ 
15 & \#ga06 & 8,266 & 5,821 & \textcolor{blue}{5.15} \\ 
16 & \#iamapreexistingcondition & 7,604 & 6,215 & \textcolor{blue}{5.41} \\ 
17 & \#himinitiative & 7,294 & 572 & \textcolor{orange}{6.04} \\ 
18 & \#news & 6,435 & 1,680 & \textcolor{blue}{5.79} \\ 
19 & \#malebreastcancer & 5,821 & 1,469 & \textcolor{orange}{6.0} \\ 
20 & \#savethetatas & 5,551 & 5,390 & \textcolor{orange}{6.11} \\ 
21 & \#giveaway & 4,861 & 1,284 & \textcolor{orange}{6.31} \\ 
22 & \#trumpcare & 4,778 & 4,331 & \textcolor{blue}{5.53} \\ 
23 & \#keepkadcyla & 3,822 & 3,064 & \textcolor{blue}{5.68} \\ 
24 & \#awareness & 3,697 & 1,369 & \textcolor{orange}{6.19} \\ 
25 & \#brca & 3,652 & 1,284 & \textcolor{blue}{5.89} \\ 
26 & \#avonrep & 3,517 & 1,620 & \textcolor{blue}{5.49} \\ 
27 & \#pink & 3,480 & 2,763 & \textcolor{orange}{6.34} \\ 
28 & \#ad & 3,458 & 1,383 & \textcolor{orange}{6.08} \\ 
29 & \#nbcf & 3,445 & 1,965 & \textcolor{orange}{6.49} \\ 
30 & \#1savetatas & 3,051 & 1,040 & \textcolor{orange}{6.51} \\ 
31 & \#worldcancerday & 2,936 & 2,430 & \textcolor{orange}{6.05} \\ 
32 & \#exercise & 2,740 & 1,492 & \textcolor{blue}{5.71} \\ 
33 & \#thinkpink & 2,707 & 2,209 & \textcolor{orange}{6.1} \\ 
34 & \#ahca & 2,607 & 2,403 & \textcolor{blue}{5.67} \\ 
35 & \#spas4acause & 2,585 & 1,615 & \textcolor{orange}{6.48} \\ 
36 & \#bcam & 2,555 & 1,961 & \textcolor{orange}{6.14} \\ 
37 & \#thegoodlie & 2,314 & 474 & \textcolor{blue}{5.63} \\ 
38 & \#healthcare & 2,261 & 1,396 & \textcolor{blue}{5.85} \\ 
39 & \#obamacare & 2,240 & 2,059 & \textcolor{orange}{6.2} \\ 
40 & \#pinkribbon & 2,201 & 1,104 & \textcolor{blue}{5.9} \\ 
41 & \#nfl & 2,188 & 647 & \textcolor{orange}{6.13} \\ 
42 & \#oncology & 2,188 & 762 & \textcolor{blue}{5.85} \\ 
43 & \#unitedbyher & 2,117 & 602 & \textcolor{orange}{6.1} \\ 
44 & \#sabcs16 & 2,104 & 828 & \textcolor{blue}{5.67} \\ 
45 & \#cnndebatenight & 2,097 & 2,029 & \textcolor{blue}{5.9} \\ 
46 & \#women & 2,078 & 1,231 & \textcolor{blue}{5.85} \\ 
47 & \#nyfw & 2,060 & 1,886 & \textcolor{orange}{6.11} \\ 
48 & \#donate & 2,016 & 1,279 & \textcolor{blue}{5.76} \\ 
49 & \#pinkout & 1,946 & 1,778 & \textcolor{orange}{6.12} \\ 
50 & \#ai & 1,937 & 1,241 & \textcolor{orange}{6.03} \\ 
\hline* & \textbf{Total} & 462,192 & 155,218 & \textcolor{darkgray}{5.96} \\ 
 \hline 

\hline 
\end{tabular}
\quad \quad
\hskip-.35cm
\begin{tabular}{|c|c|c|c|c|}
\cline{1-5}
\multicolumn{5}{|c|}{  \normalsize{\textbf{Top Hashtags(\#): Breast Cancer Patient Sample}}}\\ \hline \hline
Rank & Term & Tweets  & Users & $h_{\textnormal{avg}}$ \\ 
\hline 
1 & \#cancer & 2,063 & 239 & \textcolor{blue}{5.76} \\ 
2 & \#bcsm & 1,220 & 61 & \textcolor{orange}{5.92} \\ 
3 & \#lymphedema & 680 & 12 & \textcolor{orange}{5.93} \\ 
4 & \#breastcancer & 568 & 112 & \textcolor{orange}{5.84} \\ 
5 & \#aca & 469 & 88 & \textcolor{blue}{5.69} \\ 
6 & \#trumpcare & 168 & 70 & \textcolor{blue}{5.39} \\ 
7 & \#ahca & 165 & 45 & \textcolor{blue}{5.4} \\ 
8 & \#amsm & 165 & 25 & \textcolor{blue}{5.61} \\ 
9 & \#metastatic & 161 & 17 & \textcolor{orange}{5.92} \\ 
10 & \#malebreastcancer & 155 & 21 & \textcolor{orange}{5.94} \\ 
11 & \#worldcancerday & 134 & 54 & \textcolor{orange}{5.94} \\ 
12 & \#obamacare & 132 & 42 & \textcolor{blue}{5.77} \\ 
13 & \#saveaca & 115 & 47 & \textcolor{orange}{5.85} \\ 
14 & \#bccww & 112 & 24 & \textcolor{blue}{5.77} \\ 
15 & \#lcsm & 108 & 14 & \textcolor{orange}{5.89} \\ 
16 & \#survivor & 92 & 33 & \textcolor{orange}{5.83} \\ 
17 & \#protectourcare & 91 & 37 & \textcolor{blue}{5.75} \\ 
18 & \#iamapreexistingcondition & 82 & 55 & \textcolor{blue}{5.63} \\ 
19 & \#breast & 79 & 24 & \textcolor{orange}{6.2} \\ 
20 & \#breastcancerrealitycheck & 64 & 17 & \textcolor{blue}{5.66} \\ 
21 & \#breastcancerawarenessmonth & 62 & 41 & \textcolor{orange}{6.04} \\ 
22 & \#healthcare & 62 & 32 & \textcolor{blue}{5.44} \\ 
23 & \#kissthis4mbc & 61 & 14 & \textcolor{orange}{6.13} \\ 
24 & \#mbc & 59 & 21 & \textcolor{blue}{5.69} \\ 
25 & \#cancersucks & 57 & 34 & \textcolor{blue}{5.75} \\ 
26 & \#oncology & 54 & 11 & \textcolor{orange}{5.84} \\ 
27 & \#maga & 53 & 34 & \textcolor{blue}{5.38} \\ 
28 & \#trump & 53 & 33 & \textcolor{blue}{5.09} \\ 
29 & \#immunotherapy & 52 & 18 & \textcolor{orange}{5.81} \\ 
30 & \#clinicaltrials & 51 & 12 & \textcolor{orange}{6.13} \\ 
31 & \#acaworks & 47 & 11 & \textcolor{blue}{5.72} \\ 
32 & \#research & 46 & 17 & \textcolor{orange}{6.02} \\ 
33 & \#breastcancerawareness & 45 & 36 & \textcolor{orange}{5.89} \\ 
34 & \#f***cancer & 44 & 21 & \textcolor{orange}{5.88} \\ 
35 & \#nhs & 42 & 16 & \textcolor{blue}{5.62} \\ 
36 & \#brca & 42 & 17 & \textcolor{orange}{5.81} \\ 
37 & \#gop & 41 & 18 & \textcolor{blue}{5.28} \\ 
38 & \#metastaticbc & 41 & 19 & \textcolor{orange}{5.9} \\ 
39 & \#idrivefor & 40 & 19 & \textcolor{orange}{6.27} \\ 
40 & \#grahamcassidy & 40 & 23 & \textcolor{blue}{5.3} \\ 
41 & \#mbcproject & 39 & 11 & \textcolor{orange}{6.28} \\ 
42 & \#health & 38 & 24 & \textcolor{orange}{6.0} \\ 
43 & \#gyncsm & 37 & 10 & \textcolor{orange}{6.12} \\ 
44 & \#sabcs16 & 36 & 12 & \textcolor{blue}{5.75} \\ 
45 & \#endcancer & 35 & 13 & \textcolor{orange}{6.08} \\ 
46 & \#wecanican & 34 & 10 & \textcolor{orange}{5.83} \\ 
47 & \#savebeth & 34 & 10 & \textcolor{orange}{5.88} \\ 
48 & \#cancermoonshot & 31 & 18 & \textcolor{orange}{6.03} \\ 
49 & \#moreformbc & 30 & 16 & \textcolor{orange}{6.17} \\ 
50 & \#resist & 30 & 21 & \textcolor{blue}{5.52} \\ 
\hline* & \textbf{Total} & 8,159 & 398 & \textcolor{darkgray}{5.81} \\ 
 \hline 

\hline 
\end{tabular}
\caption{ {\textbf{50 Most Frequently Tweeted Hashtags:} A table of the most frequently tweeted hashtags (\#) from all collected breast cancer tweets (left) and from sampled breast cancer patients (right).  The relative computed ambient happiness $h_{\textnormal{avg}}$ for each hashtag is colored relative to the group average (blue- negative, orange - positive).   }} 
\label{table:Hashtags}
\end{table}

\pagebreak

\section{Discussion}

 We have demonstrated the potential of using sentence classification to isolate content authored by breast cancer patients and survivors.  Our novel, multi-step sifting algorithm helped us differentiate topics relevant to patients and compare their sentiments to the global online discussion.  The hedonometric comparison of frequent hashtags helped identify prominent topics  how their sentiments differed.  This shows the ambient happiness scores of terms and topics can provide useful information regarding comparative emotionally charged content.  This process can be applied to disciplines across health care and beyond.  
 
   Throughout 2017, Healthcare was identified as a pressing issue causing anguish and fear among the breast cancer community; especially among patients and survivors.  During this time frame, US legislation was proposed by Congress that could roll back regulations ensuring coverage for individuals with pre-existing conditions.  Many individuals identifying as current breast cancer patients/survivors expressed concerns over future treatment and potential loss of their healthcare coverage.  Twitter could provide a useful political outlet for patient populations to connect with legislators and sway political decisions.
  
  March 2017 was a relatively negative month due to discussions over American healthcare reform.  The American Congress held a vote to repeal the Affordable Care Act (ACA, also referred to as `Obamacare'), which could potentially leave many Americans without healthcare insurance, \cite{NYT_healthcare_2017-03-23}.  There was an overwhelming sense of apprehension within the `breast cancer' tweet sample.  Many patients/survivors in our diagnostic tweet sample identified their condition and how the ACA ensured coverage throughout their treatment.  
  
     This period featured a notable tweet frequency spike, comparable to the peak during breast cancer awareness month. The  burst event  peaked on March 23rd and 24th (65k, 57k tweets respectively, see Figure \ref{fig:TweetFeatures}).  During the peak, 41,983 (34\%) posts contained `care' in reference to healthcare, with a viral retweeted meme accounting for 39,183 of these mentions. The tweet read: "The group proposing to cut breast cancer screening, maternity care, and contraceptive coverage." with an embedded photo of a group of predominately male legislators, \cite{AcaTweetMeme}. The criticism referenced the absence of female representation in a decision that could deprive many of coverage for breast cancer screenings. The online community condemned the decision to repeal and replace the ACA with the proposed legislation with references to people in treatment who could `die' (n=7,923) without appropriate healthcare insurance coverage. The vote was later postponed and eventually failed, \cite{CNN_healthcare_2017-07-28}. 
     
     Public outcry likely influenced this legal outcome, demonstrating Twitter's innovative potential as a support tool for public lobbying of health benefits. Twitter can further be used to remind, motivate and change individual and population health behavior using messages of encouragement (translated to happiness) or dissatisfaction (translated to diminished happiness), for example, with memes that can have knock on social consequences when they are re-tweeted.  Furthermore, Twitter may someday be used to benchmark treatment decisions to align with expressed patient sentiments, and to make or change clinical recommendations based upon the trend histories that evolve with identifiable sources but are entirely in the public domain. 

\hspace{5mm} Analyzing the fluctuation in average word happiness as well as bursts in the frequency distributions can help identify relevant events for further investigation. These tools helped us extract themes relevant  to breast cancer patients in comparison to the global conversation.
\begin{table}[H]
\hskip3cm
\begin{tabular}{|c|l|}
\cline{1-2}
No. & Tweet \\ 
\hline
\hline

1 & i was diagnosed ... with stage 2 breast cancer ... after 4 years in remission ... \\
\hline
2 &  obamacare saved me! i have had breast cancer twice ... \\
\hline
3 & yesterday i was diagnosed with breast cancer ... \\
\hline

4 & ... i have breast cancer but i will get through this ...   \\

\hline
5 & ... i've had .. breast cancer and  ... i can't get insurance because i can't afford it \\

\hline
\hline
\end{tabular}
\caption{ {\textbf{Sampled  Predicted Diagnostic Tweets:}  A sample of key phrases from self-reported diagnostic tweets predicted from the CNN classifier with the patient relevant proportional ratio, $\alpha=1:10$.}} 
\end{table}

One area in which Twitter has traditionally fallen short for a communication medium is that of the aural dimension, such as nuances and inflections. However, Twitter now includes pictures, videos and emojis with people revealing or conveying their emotions by use of these communication methods.  It is envisaged that the aural and visual dimensions will eventually grow to complement the published text component towards a more refined understanding of feelings, attitudes and health and clinical sentiments. 

  Lack of widespread patient adoption of social media could be a limiting factor to our analysis.  A study of breast cancer patients during  2013--2014, \citet{wallner2016use}, found social media was a less prominent form of online communication (N = 2578, 12.3\%), however with the advent of smartphones and the internet of things (iot) movement, social media may influence a larger proportion of future patients.  Another finding noted that online posts were more likely to be positive about their healthcare decision experience or about survivorship. Therefore we cannot at this time concretely draw population-based assumptions from social media sampling.  Nevertheless, understanding this online patient community could serve as a valuable tool for healthcare providers and future studies should investigate current social media usage statistics across patients.
  
  Because we trained the content classifier with a relatively small corpus, the model likely over-fit on a few particular word embeddings.  For example: 'i have stage iv', `i am * survivor', `i had * cancer'. However, this is similar to the process of recursive keyword searches to gather related content. Also, the power of the CNN allows for multiple relative lingual syntax as opposed to searching for static phrases ('i have breast cancer', 'i am a survivor').  The CNN shows great promise in sifting relevant context from large sets of data.

      Other social forums for patient self reporting and discussion should be incorporated into future studies.  For example, as of 2017, \url{https://community.breastcancer.org} has built a population of over 199,000 members spanning 145,000 topics.  These tools could help connect healthcare professionals with motivated patients.  Labeled posts from patients could also help train future context models and help identify adverse symptoms shared among online social communities. 
  
Our study focused primarily on English tweets, since this was the language of our diagnostic training sample. Future studies could incorporate other languages using our proposed framework.  It would be important to also expand the API queries with translations of `breast' and `cancer'.  This could allow for a cross cultural comparison of how social media influences patients and what patients express on social media. 

 \section{Conclusion}

\hspace{5mm} We have demonstrated the potential of using context classifiers for identifying diagnostic tweets related to the experience of breast cancer patients. Our framework provides a proof of concept for integrating machine learning with natural language processing as a tool to help connect healthcare providers with patient experiences. These methods can inform the medical community to provide more personalized treatment regimens by evaluating patient satisfaction using social listening. Twitter has also been shown as a useful medium for political support of healthcare policies as well as spreading awareness. Applying these analyses across other social media platforms could provide comparably rich data-sets. For instance, Instagram has been found to contain indicative markers for depression, \citet{reece2016instagram}. Integrating these applications into our healthcare system could provide a better means of tracking iPROs across treatment regimens and over time. \\

One area in which Twitter has traditionally fallen short for a communication medium is that of the aural dimension, such as nuances and inflections. However, Twitter now includes pictures, videos, and emojis with people revealing or conveying their emotions by use of these communication methods. With augmented reality, virtual reality, and even chatbot interfaces, it is envisaged that the aural and visual dimensions will eventually grow to complement the published text component towards a more refined understanding of feelings, attitudes and health and clinical sentiments. 

Follow-on studies to our work could be intended to further develop these models and apply them to larger streams of data. Online crowd sourcing tools, like Amazon's Mechanical Turk, implemented in, \citet{dodds2014human}, can help compile larger sets of human validated labels to improve context classifiers. These methods can also be integrated into delivering online outreach surveys as another tool for validating healthcare providers.  Future models, trained on several thousand labeled tweets for various real world applications should be explored. Invisible patient- reported outcomes should be further investigated via sentiment and context analyses for a better understanding of how to integrate the internet of things with healthcare. 

  Twitter has become a powerful platform for amplifying political voices of individuals.  The response of the online breast cancer community to the American Healthcare Act as a replacement to the Affordable Care Act was largely negative due to concerns over loss of coverage.  A widespread negative public reaction helped influence this political result.  Social media opinion mining could present as a powerful tool for legislators to connect with and learn from their constituents. This can lead to positive impacts on population health and societal well-being.

\section{Acknowledgments}

\hspace{5mm} The authors wish to acknowledge the Vermont Advanced Computing Core, which is supported by NASA (NNX-08AO96G) at the University of Vermont which provided High Performance Computing resources that contributed to the research results reported within this poster. EMC was supported by the Vermont Complex Systems Center.  CMD and PSD were supported by an NSF BIGDATA grant IIS-1447634.

\clearpage

\section*{Supplementary materials}

\section*{Appendix I:  Raw `Cancer' Twitter Data Overview }

 \begin{figure}[H]
  \hskip-1cm
  \includegraphics[scale=.2]{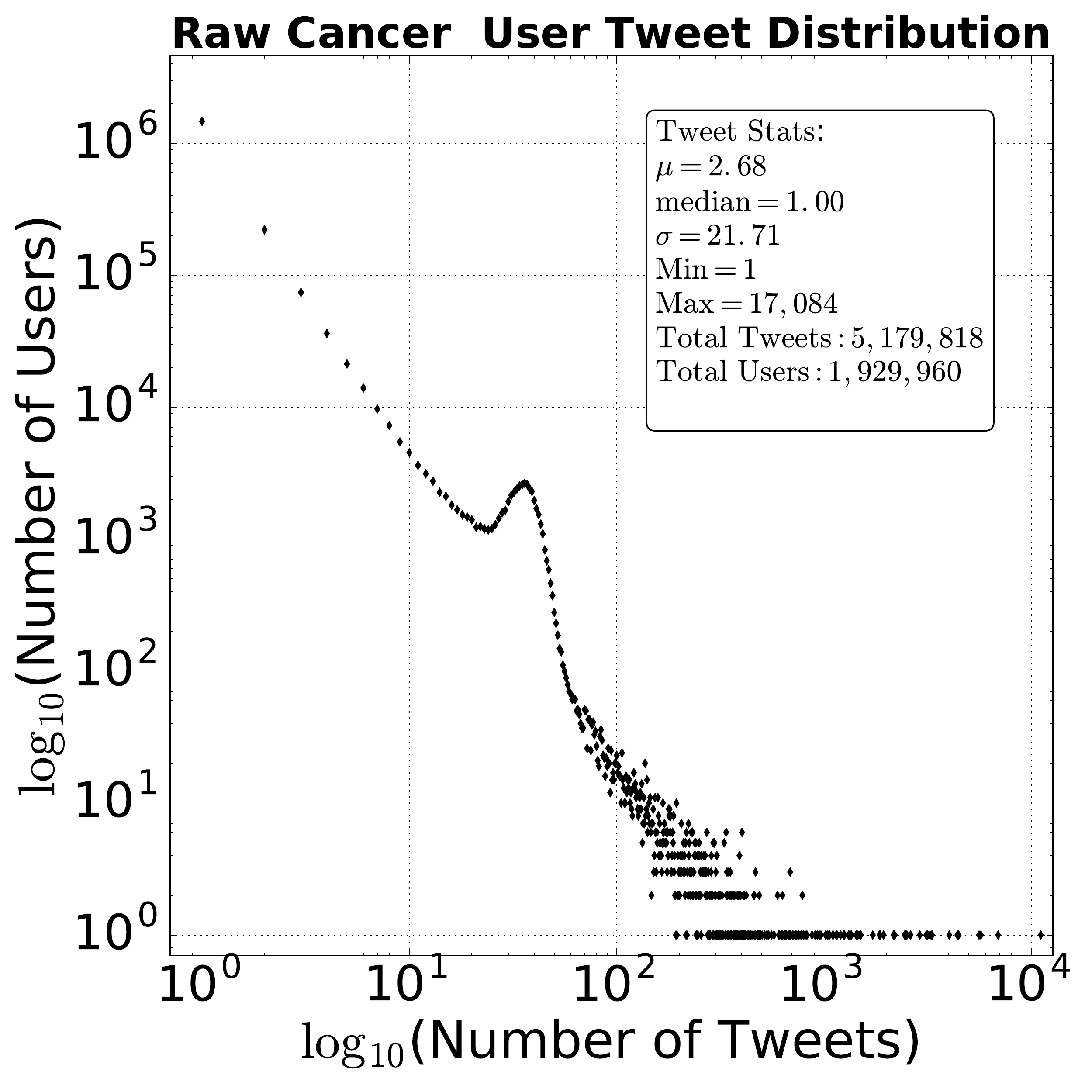}
 \includegraphics[scale=.3]{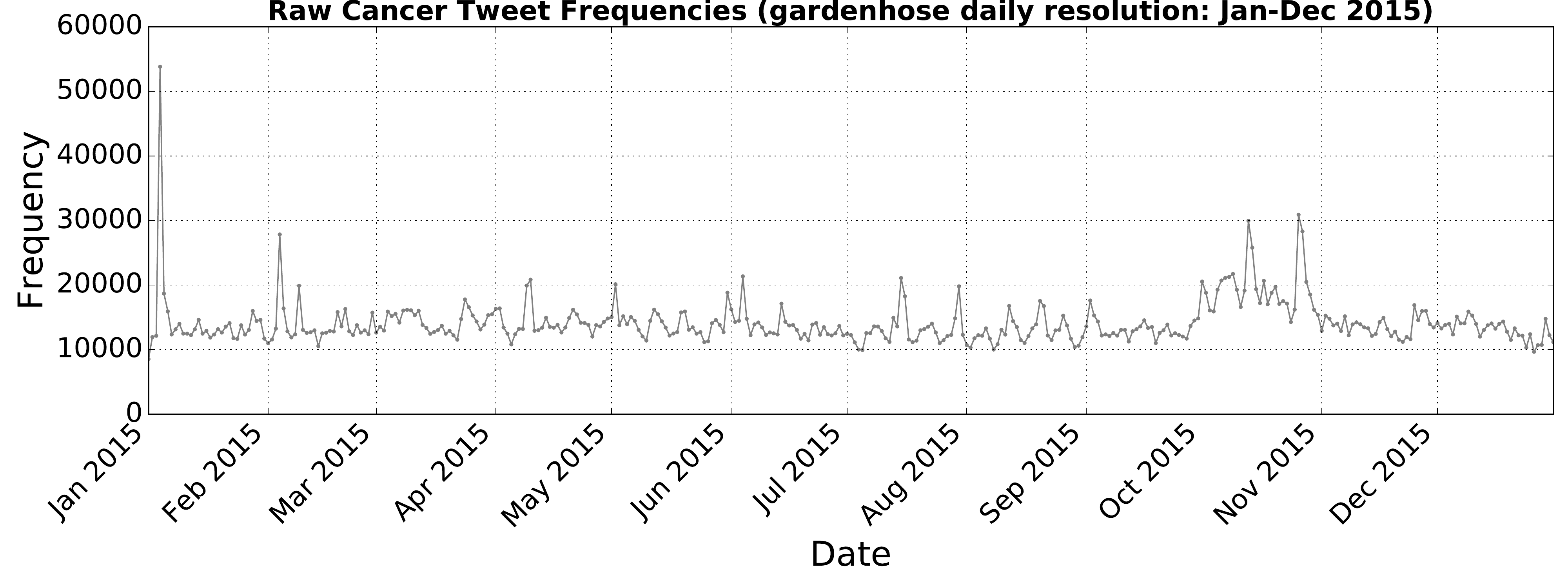}
 \caption*{Figure A1-1: A Frequency time-series of raw `cancer' tweets collected, binned by day.  This sample was compiled from a 10\% random sample of Twitter, the `Gardenhose' feed.   (left) The distribution of tweets per given user is plotted on a log axis. The tail tends to be high frequency automated accounts, some of which provide daily updates on horoscope information, or about news related to cancer.  The kink in the center is also abnormal and could be representative of other classes of automation.  This shows the necessity to sift irrelevant tweets using combinations of keyword removal and content classifiers. }
\label{fig:TweetFeatures_RawCancer}
 \end{figure}

 \begin{figure}[H]
  \hskip-1cm
  \includegraphics[scale=.2]{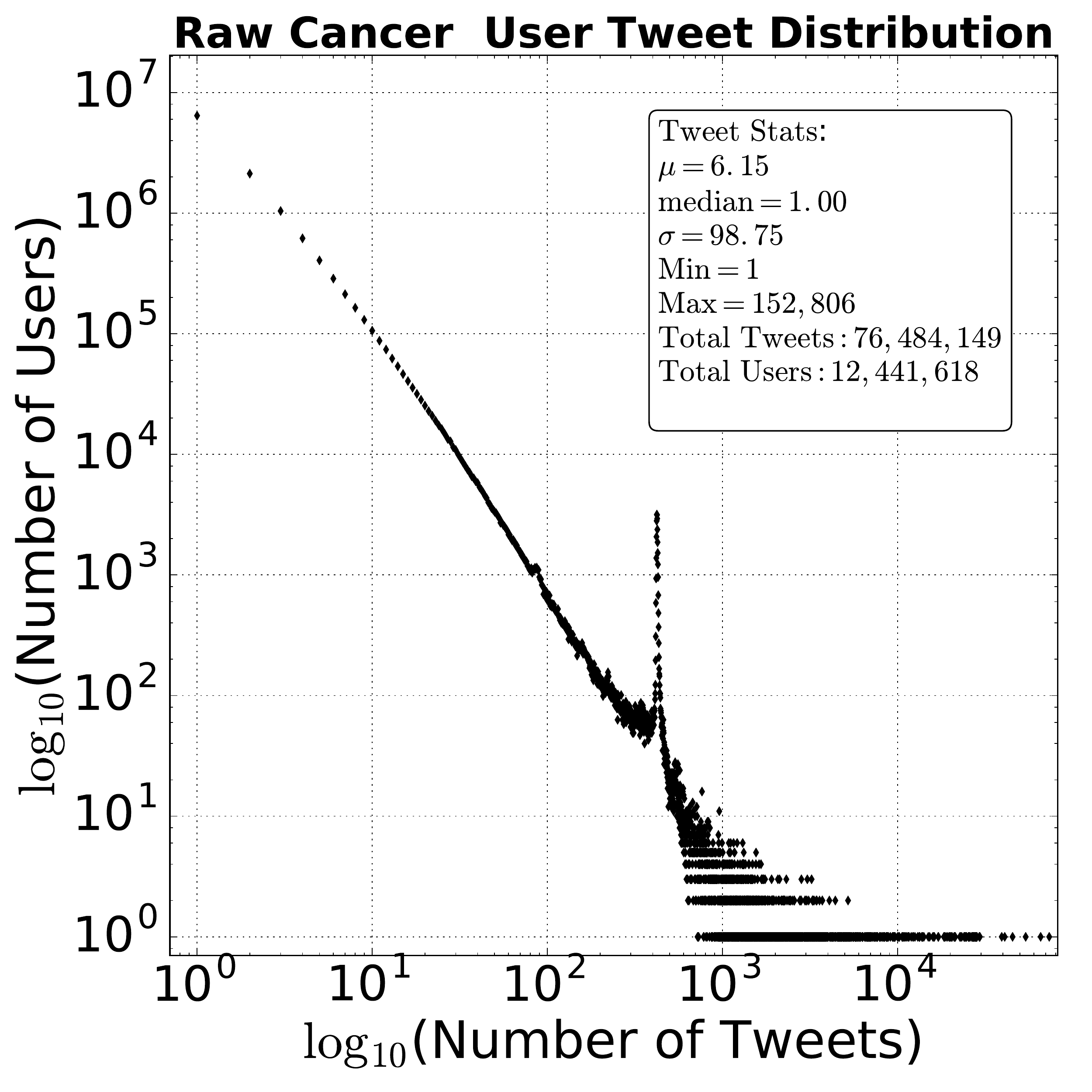}
 \includegraphics[scale=.3]{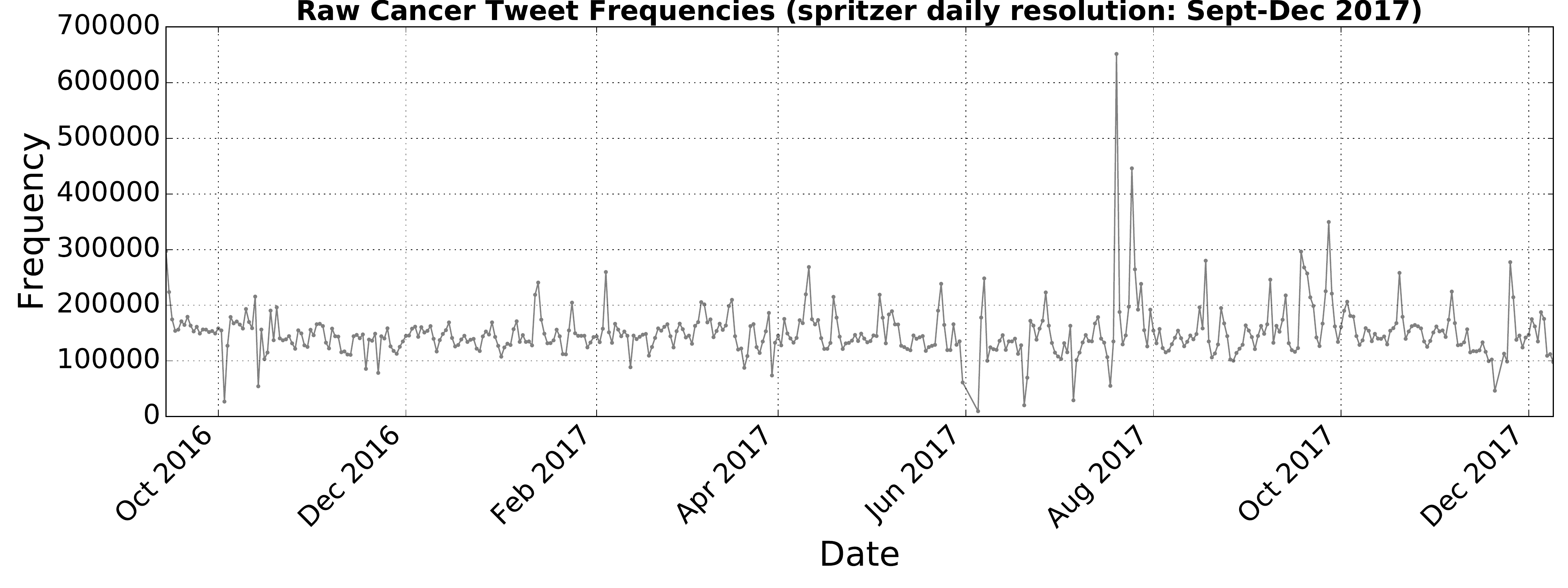}
 \caption*{Figure A1-2:  Another frequency time-series of raw `cancer' tweets collected, binned by day.  This sample was compiled from a 1\% random sample of Twitter, the `Spritzer' feed concentrated on keyword `cancer', during the same time interval as Figure \ref{fig:TweetFeatures}.   We collected over 76 million tweets, which accounted for approximately 65.2\% of all tweets mentioning `cancer' while the data stream was active (i.e., not accounting for power/network outages).    The kink that was visible in the previous figure seems to moved outward by almost a factor of 10, since this is a much larger sample of `cancer' tweets  (10\% versus $\approx$ 65\%).  This serves as a comparison to the tweets collected using  keywords `breast' and `cancer' and to raw `cancer' tweets collected from the Gardenhose feed.}
\label{fig:TweetFeatures_RawCancerSpritzer}
 \end{figure}

 \pagebreak
 
 \section*{Appendix II: Calculating the Tweet Sampling Proportion}

\hspace{5mm} There are three types of endpoints to access data from Twitter.  The `spritzer' (1\%) and `gardenhose' (10\%) endpoints were both implemented to collect publicly posted relevant data for our analysis. The third type of endpoint is the `Firehose' feed, a full 100\% sample, which can be purchased via subscription from Twitter.   This was unnecessary for our analysis, since our set of keywords yielded a high proportion of the true tweet sample.  We quantified the sampled proportion of tweets using overflow statistics provided by Twitter. These `limit tweets', $L$, issue a timestamp along with the approximate number of posts withheld from our collected sample, $T_s$.  The sampling percentage, $\tilde{\rho}_s$, of keyword tweets is approximated as the collected tweet total, $\vert T_s \vert$ , as a proportion of itself combined with the sum of the limit counts, each $\ell \in L$:

\begin{equation}
\displaystyle \tilde{\rho}_s = \dfrac{\displaystyle \vert T_s  \vert }{ \displaystyle \vert T_s  \vert  + \sum_{\ell \in L} \ell} = \dfrac{\text{total collected tweets}}{\text{total collected tweets + overflow limit sum}} \approx \text{sampling proportion} 
\end{equation}

By the end of 2017, Twitter was accumulating an average of 500 million tweets per day, \citet{TwitterUsageStats}.  Our topics were relatively specific, which allowed us to collect a large sample of tweets.  For the singular search term, `cancer',  the keyword sampled proportion, $\tilde{\rho}_s$, was approximately 65.21\% with a sample of 89.2 million tweets.  Our separate Twitter spritzer feed searching for keywords `breast AND cancer` OR `lymphedema'  rarely surpassed the 1\% limit.  We calculated a 96.1\% sampling proportion while our stream was active (i.e. not accounting for network or power outages).  We present the daily overflow limit counts of tweets not appearing in our data-set, and the approximation of the sampling size in Figure A2. 
 \begin{figure}[H]
 \centering
   \includegraphics[scale=.4]{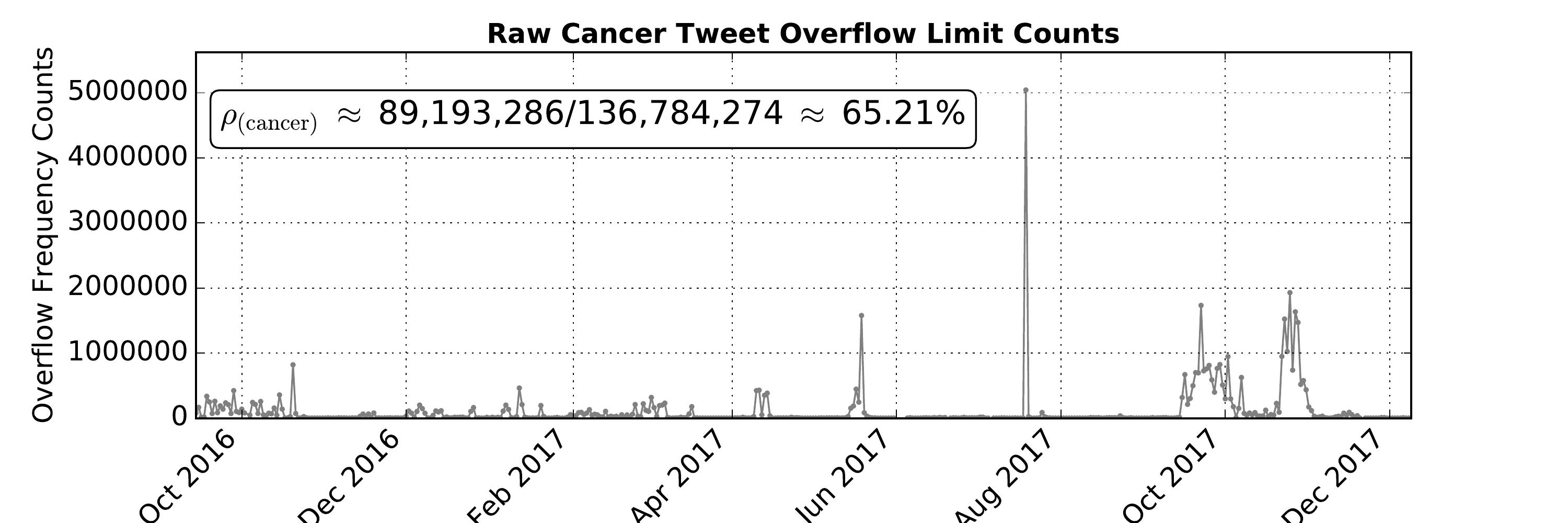}
   \includegraphics[scale=.4]{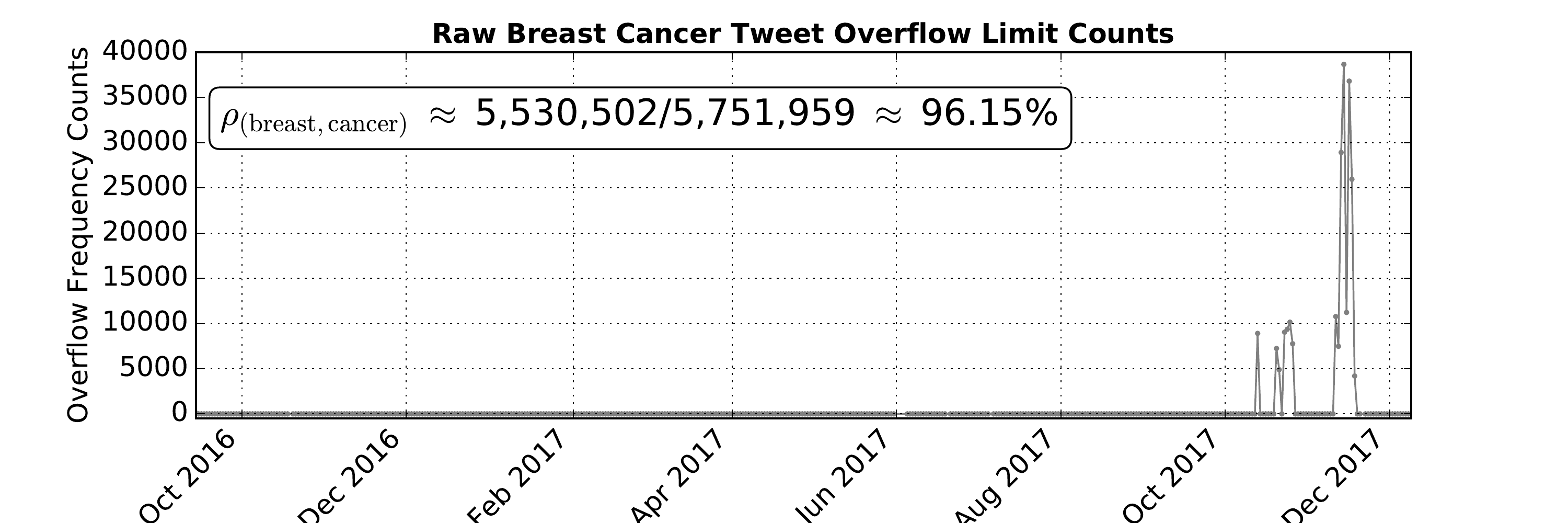}
\caption*{\textbf{Figure A2:} Overflow limit statistics, plotted per day for both the cancer and breast cancer Twitter feeds with the corresponding approximation of the sampling proportion over the study time frame.   }
\label{fig:OverflowLimits}
 \end{figure}

\vspace{8cm}

 \pagebreak
 
  \pagestyle{empty}    \section*{Appendix III:  Interpreting Word Shift Graphs}
  \vskip-.2cm
 \hspace{5mm} Word shift graphs are essential tools for analyzing which terms are affecting the computed average happiness scores between two text distributions, \citet{reagan2015benchmarking}.  The reference word distribution, $T_{\textnormal{ref}}$, serves as a lingual basis to compare with another text, $T_{\textnormal{comp}}$.  The top 50 words causing the shift in computed word happiness are displayed along with their relative weight.    The arrows ($\uparrow, \downarrow$) next to each word mark  an increase or decrease in the word's frequency.  The $+$,$-$, symbols indicate whether the word contributes positively or negatively to the shift in computed average word happiness.

 In Figure A3,  word shift graphs compare tweets mentioning `breast' `cancer' and a random 10\% `Gardenhose' sample of non filtered tweets.  On the left,  `breast',`cancer'  tweets were slightly less positive due to an increase in negative words like `fight', `battle', `risk', and `lost'.  These distributions had similar average happiness scores, which was in part due to the relatively more positive words  `women', mom', `raise', `awareness', `save', `support', and `survivor'.  The word shift on the right compares breast cancer patient tweets to non filtered tweets.  These were more negative ($h_{\textnormal{avg}}$ = 5.78 v. 6.01) due a relative increase in words like `fighting', `surgery', `against', `dying', `sick', `killing', `radiation', and `hospital'.  This tool helped identify words that signal emotional themes and allow us to extract content from large corpora, and identify thematic emotional topics within the data.

 \begin{figure}[H]
 \centering
 \hskip-1cm
   \includegraphics[scale=.38]{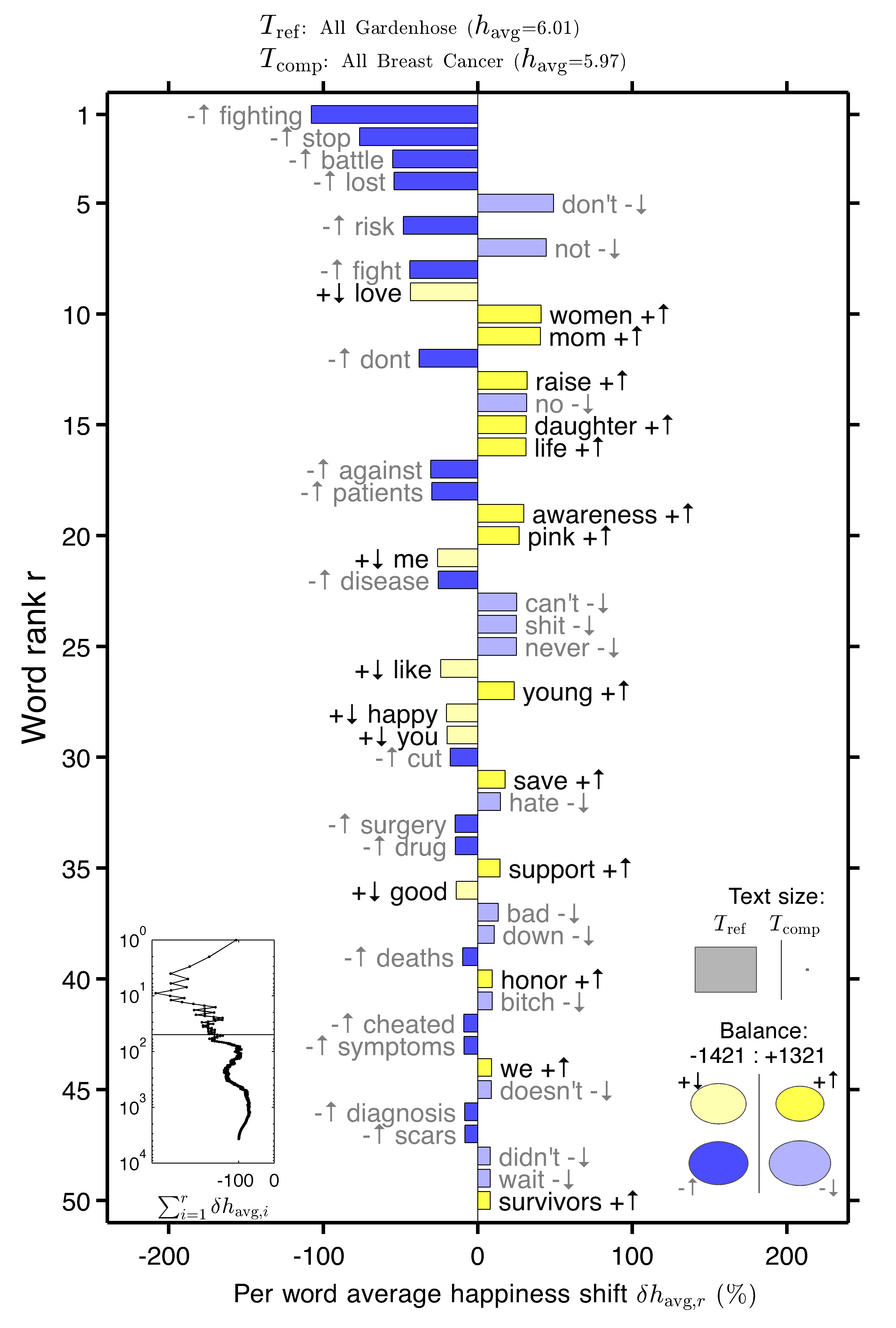}
  \includegraphics[scale=.38]{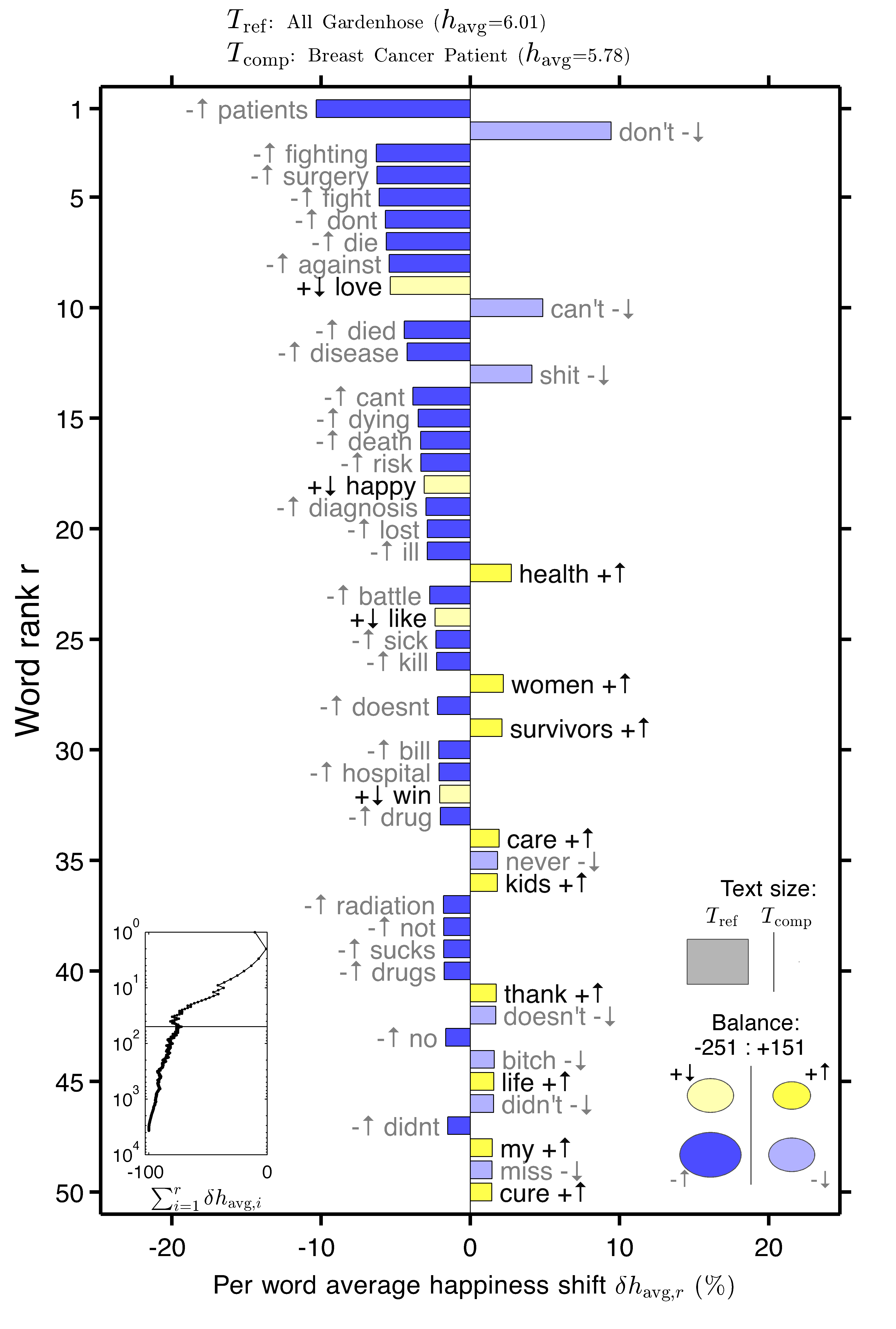}
\caption*{\footnotesize{\textbf{Figure A3:} (Left) A word shift graph comparing tweets collected mentioning breast cancer, $T_{\textnormal{comp}}$, to a random unfiltered reference sample of tweets along the same time period.  Breast cancer tweets were slightly less positive ($h_{\textnormal{avg}} = 5.97$ v. $6.01$) due to an increase in negative words `fight(ing)', `stop' , `battle', `lost', and `risk'.  This set of tweets featured a relative increase in positive words `women', `mom', `daughter', `awareness', `pink', `save', `support', and `survivors', which are referencing aspects of breast cancer awareness, support,  and the experiences of survivors and patients. (Right) A word shift graph comparing breast cancer patient tweets to the unfiltered sample.  These were more negative ($h_{\textnormal{avg}}$ = 5.78 v. 6.01) due to a relative increase in negative words such as `dying', `sick', `killing', `radiation', and `hospital'   among other terms similar to the figure on the left. }}
\label{fig:WordShiftRef}
 \end{figure}
 
 \pagebreak

 \section*{ Appendix IV: Sentence Classification Methodology} 
 
 We built the vocabulary corpus for the logistic model by tokenizing the annotated set of patient tweets by word, removing punctuation, and lowercasing all text.  We also included patient unrelated `cancer'  tweets collected as a frame of reference to train the classifier.  This set of tweets was not annotated, so we made the assumption that tweets not validated by, \citet{crannell2016pattern}  were patient unrelated.  The proportion, $\alpha$, of unrelated to related  tweets has a profound effect on the vocabulary of the logistic model, so we experimented with various ranges of $\alpha$ and settled on a 1:10 ratio of patient related to unrelated tweets.  We then applied the tf-idf statistic to build the binary classification logistic model.

 The Tensorflow open source machine learning library has previously shown great promise when applied to NLP benchmark data-sets, \citet{kim2014convolutional} .  The CNN loosely works by implementing a filter, called convolution functions, across various subregions of the feature landscape, \citet{johnson2015semi, UnderstandingCNNs}, in this case the tweet vocabulary.  The model tests the robustness of different word embeddings (e.g., phrases) by randomly removing filtered pieces during optimization to find the best predictive terms over the course of training.  We divided the input labeled data into training and evaluation to successively test for the best word embedding predictors.  The trained model can then be applied for binary classification of text content.      
 
 \vspace{12cm}
 
 \pagebreak

%
 %\begin{textblock*}{16cm}(.13\textwidth,-2.5cm)
%\large{Appendix IV: Hashtag Table Sorted by Average Word Happiness}
%\end{textblock*}

 \section*{ Appendix V: Hashtag Table Sorted by Average Word Happiness} 

     \pagestyle{empty}  \begin{table}[H]
\begin{tabular}{|c|c|c|c|c|}
\cline{1-5}
\multicolumn{5}{|c|}{  \normalsize{\textbf{Top Hashtags(\#): All Breast Cancer}}}\\ \hline \hline
Rank & Term & Tweets  & Users & $h_{\textnormal{avg}}$ \\ 
\hline 
1 & \#1savetatas & 3,051 & 1,040 & \textcolor{orange}{6.51} \\ 
2 & \#nbcf & 3,445 & 1,965 & \textcolor{orange}{6.49} \\ 
3 & \#spas4acause & 2,585 & 1,615 & \textcolor{orange}{6.48} \\ 
4 & \#pink & 3,480 & 2,763 & \textcolor{orange}{6.34} \\ 
5 & \#giveaway & 4,861 & 1,284 & \textcolor{orange}{6.31} \\ 
6 & \#obamacare & 2,240 & 2,059 & \textcolor{orange}{6.2} \\ 
7 & \#awareness & 3,697 & 1,369 & \textcolor{orange}{6.19} \\ 
8 & \#twibbon & 16,809 & 14,332 & \textcolor{orange}{6.18} \\ 
9 & \#bcam & 2,555 & 1,961 & \textcolor{orange}{6.14} \\ 
10 & \#breastcancerawareness & 13,429 & 8,820 & \textcolor{orange}{6.13} \\ 
11 & \#nfl & 2,188 & 647 & \textcolor{orange}{6.13} \\ 
12 & \#pinkout & 1,946 & 1,778 & \textcolor{orange}{6.12} \\ 
13 & \#savethetatas & 5,551 & 5,390 & \textcolor{orange}{6.11} \\ 
14 & \#nyfw & 2,060 & 1,886 & \textcolor{orange}{6.11} \\ 
15 & \#thinkpink & 2,707 & 2,209 & \textcolor{orange}{6.1} \\ 
16 & \#unitedbyher & 2,117 & 602 & \textcolor{orange}{6.1} \\ 
17 & \#ad & 3,458 & 1,383 & \textcolor{orange}{6.08} \\ 
18 & \#idrivefor & 13,562 & 8,331 & \textcolor{orange}{6.06} \\ 
19 & \#breastcancerawarenessmonth & 20,961 & 13,491 & \textcolor{orange}{6.06} \\ 
20 & \#aca & 8,903 & 8,105 & \textcolor{orange}{6.05} \\ 
21 & \#worldcancerday & 2,936 & 2,430 & \textcolor{orange}{6.05} \\ 
22 & \#himinitiative & 7,294 & 572 & \textcolor{orange}{6.04} \\ 
23 & \#ai & 1,937 & 1,241 & \textcolor{orange}{6.03} \\ 
24 & \#walk & 9,344 & 246 & \textcolor{orange}{6.0} \\ 
25 & \#malebreastcancer & 5,821 & 1,469 & \textcolor{orange}{6.0} \\ 
26 & \#breast & 35,544 & 11,115 & \textcolor{orange}{6.0} \\ 
27 & \#survivor & 14,500 & 1,107 & \textcolor{orange}{5.98} \\ 
28 & \#breastcancer & 66,400 & 22,247 & \textcolor{orange}{5.97} \\ 
29 & \#research & 1,912 & 1,634 & \textcolor{orange}{5.96} \\ 
30 & \#bcsm & 14,955 & 4,644 & \textcolor{blue}{5.95} \\ 
31 & \#cancer & 67,111 & 23,171 & \textcolor{blue}{5.92} \\ 
32 & \#cnndebatenight & 2,097 & 2,029 & \textcolor{blue}{5.9} \\ 
33 & \#pinkribbon & 2,201 & 1,104 & \textcolor{blue}{5.9} \\ 
34 & \#brca & 3,652 & 1,284 & \textcolor{blue}{5.89} \\ 
35 & \#lymphedema & 13,263 & 2,274 & \textcolor{blue}{5.88} \\ 
36 & \#women & 2,078 & 1,231 & \textcolor{blue}{5.85} \\ 
37 & \#healthcare & 2,261 & 1,396 & \textcolor{blue}{5.85} \\ 
38 & \#oncology & 2,188 & 762 & \textcolor{blue}{5.85} \\ 
39 & \#health & 17,484 & 5,696 & \textcolor{blue}{5.82} \\ 
40 & \#news & 6,435 & 1,680 & \textcolor{blue}{5.79} \\ 
41 & \#nobraday & 23,406 & 16,785 & \textcolor{blue}{5.76} \\ 
42 & \#donate & 2,016 & 1,279 & \textcolor{blue}{5.76} \\ 
43 & \#exercise & 2,740 & 1,492 & \textcolor{blue}{5.71} \\ 
44 & \#keepkadcyla & 3,822 & 3,064 & \textcolor{blue}{5.68} \\ 
45 & \#sabcs16 & 2,104 & 828 & \textcolor{blue}{5.67} \\ 
46 & \#ahca & 2,607 & 2,403 & \textcolor{blue}{5.67} \\ 
47 & \#thegoodlie & 2,314 & 474 & \textcolor{blue}{5.63} \\ 
48 & \#trumpcare & 4,778 & 4,331 & \textcolor{blue}{5.53} \\ 
49 & \#avonrep & 3,517 & 1,620 & \textcolor{blue}{5.49} \\ 
50 & \#iamapreexistingcondition & 7,604 & 6,215 & \textcolor{blue}{5.41} \\ 
\hline* & \textbf{Total} & 462,192 & 155,218 & \textcolor{darkgray}{5.96} \\ 
 \hline 

\hline 
\end{tabular}
\quad \quad
\begin{tabular}{|c|c|c|c|c|}
\cline{1-5}
\multicolumn{5}{|c|}{  \normalsize{\textbf{Top Hashtags(\#): Breast Cancer Patient Sample}}}\\ \hline \hline
Rank & Term & Tweets  & Users & $h_{\textnormal{avg}}$ \\ 
\hline 
1 & \#crucialcatch & 107 & 3 & \textcolor{orange}{6.58} \\ 
2 & \#mbcproject & 39 & 11 & \textcolor{orange}{6.28} \\ 
3 & \#idrivefor & 40 & 19 & \textcolor{orange}{6.27} \\ 
4 & \#breast & 79 & 24 & \textcolor{orange}{6.2} \\ 
5 & \#childhoodcancer & 42 & 9 & \textcolor{orange}{6.14} \\ 
6 & \#clinicaltrials & 51 & 12 & \textcolor{orange}{6.13} \\ 
7 & \#kissthis4mbc & 61 & 14 & \textcolor{orange}{6.13} \\ 
8 & \#breastcancerawarenessmonth & 62 & 41 & \textcolor{orange}{6.04} \\ 
9 & \#research & 46 & 17 & \textcolor{orange}{6.02} \\ 
10 & \#lifeofafourthstager & 41 & 3 & \textcolor{orange}{6.0} \\ 
11 & \#aacr17 & 51 & 7 & \textcolor{orange}{5.99} \\ 
12 & \#curechat & 70 & 3 & \textcolor{orange}{5.97} \\ 
13 & \#mylymphedemalife & 170 & 4 & \textcolor{orange}{5.95} \\ 
14 & \#worldcancerday & 134 & 54 & \textcolor{orange}{5.94} \\ 
15 & \#malebreastcancer & 155 & 21 & \textcolor{orange}{5.94} \\ 
16 & \#lymphedema & 680 & 12 & \textcolor{orange}{5.93} \\ 
17 & \#metastatic & 161 & 17 & \textcolor{orange}{5.92} \\ 
18 & \#bcsm & 1,220 & 61 & \textcolor{orange}{5.92} \\ 
19 & \#metastaticbc & 41 & 19 & \textcolor{orange}{5.9} \\ 
20 & \#lcsm & 108 & 14 & \textcolor{orange}{5.89} \\ 
21 & \#breastcancerawareness & 45 & 36 & \textcolor{orange}{5.89} \\ 
22 & \#f***cancer & 44 & 21 & \textcolor{orange}{5.88} \\ 
23 & \#cpat17 & 61 & 3 & \textcolor{orange}{5.87} \\ 
24 & \#saveaca & 115 & 47 & \textcolor{orange}{5.85} \\ 
25 & \#breastcancer & 568 & 112 & \textcolor{orange}{5.84} \\ 
26 & \#oncology & 54 & 11 & \textcolor{orange}{5.84} \\ 
27 & \#survivor & 92 & 33 & \textcolor{orange}{5.83} \\ 
28 & \#immunotherapy & 52 & 18 & \textcolor{orange}{5.81} \\ 
29 & \#brca & 42 & 17 & \textcolor{orange}{5.81} \\ 
30 & \#obamacare & 132 & 42 & \textcolor{blue}{5.77} \\ 
31 & \#bccww & 112 & 24 & \textcolor{blue}{5.77} \\ 
32 & \#cancer & 2,063 & 239 & \textcolor{blue}{5.76} \\ 
33 & \#protectourcare & 91 & 37 & \textcolor{blue}{5.75} \\ 
34 & \#cancersucks & 57 & 34 & \textcolor{blue}{5.75} \\ 
35 & \#acaworks & 47 & 11 & \textcolor{blue}{5.72} \\ 
36 & \#aca & 469 & 88 & \textcolor{blue}{5.69} \\ 
37 & \#mbc & 59 & 21 & \textcolor{blue}{5.69} \\ 
38 & \#breastcancerrealitycheck & 64 & 17 & \textcolor{blue}{5.66} \\ 
39 & \#chokecancer & 303 & 1 & \textcolor{blue}{5.64} \\ 
40 & \#iamapreexistingcondition & 82 & 55 & \textcolor{blue}{5.63} \\ 
41 & \#nhs & 42 & 16 & \textcolor{blue}{5.62} \\ 
42 & \#amsm & 165 & 25 & \textcolor{blue}{5.61} \\ 
43 & \#projectpinkblue & 178 & 1 & \textcolor{blue}{5.56} \\ 
44 & \#healthcare & 62 & 32 & \textcolor{blue}{5.44} \\ 
45 & \#ahca & 165 & 45 & \textcolor{blue}{5.4} \\ 
46 & \#trumpcare & 168 & 70 & \textcolor{blue}{5.39} \\ 
47 & \#maga & 53 & 34 & \textcolor{blue}{5.38} \\ 
48 & \#grahamcassidy & 40 & 23 & \textcolor{blue}{5.3} \\ 
49 & \#stageivneedsmore & 51 & 8 & \textcolor{blue}{5.29} \\ 
50 & \#gop & 41 & 18 & \textcolor{blue}{5.28} \\ 
\hline* & \textbf{Total} & 8,928 & 396 & \textcolor{darkgray}{5.8} \\ 
 \hline 

\hline 
\end{tabular}
\caption{ {\textbf{50 Most Frequently Tweeted Hashtags:} A table of the most frequently tweeted hashtags (\#) from all collected breast cancer tweets (left) and from sampled breast cancer patients (right).  The relative computed average happiness $h_{\textnormal{avg}}$ for each tag is colored relative to the group average (blue- negative, orange - positive). This version is sorted by computed word happiness.   }} 
\label{table:Hashtags}
\end{table}

\pagebreak

 \begin{figure}[H]
 \centering
 \hskip-1cm
   \includegraphics[width=.99\linewidth]{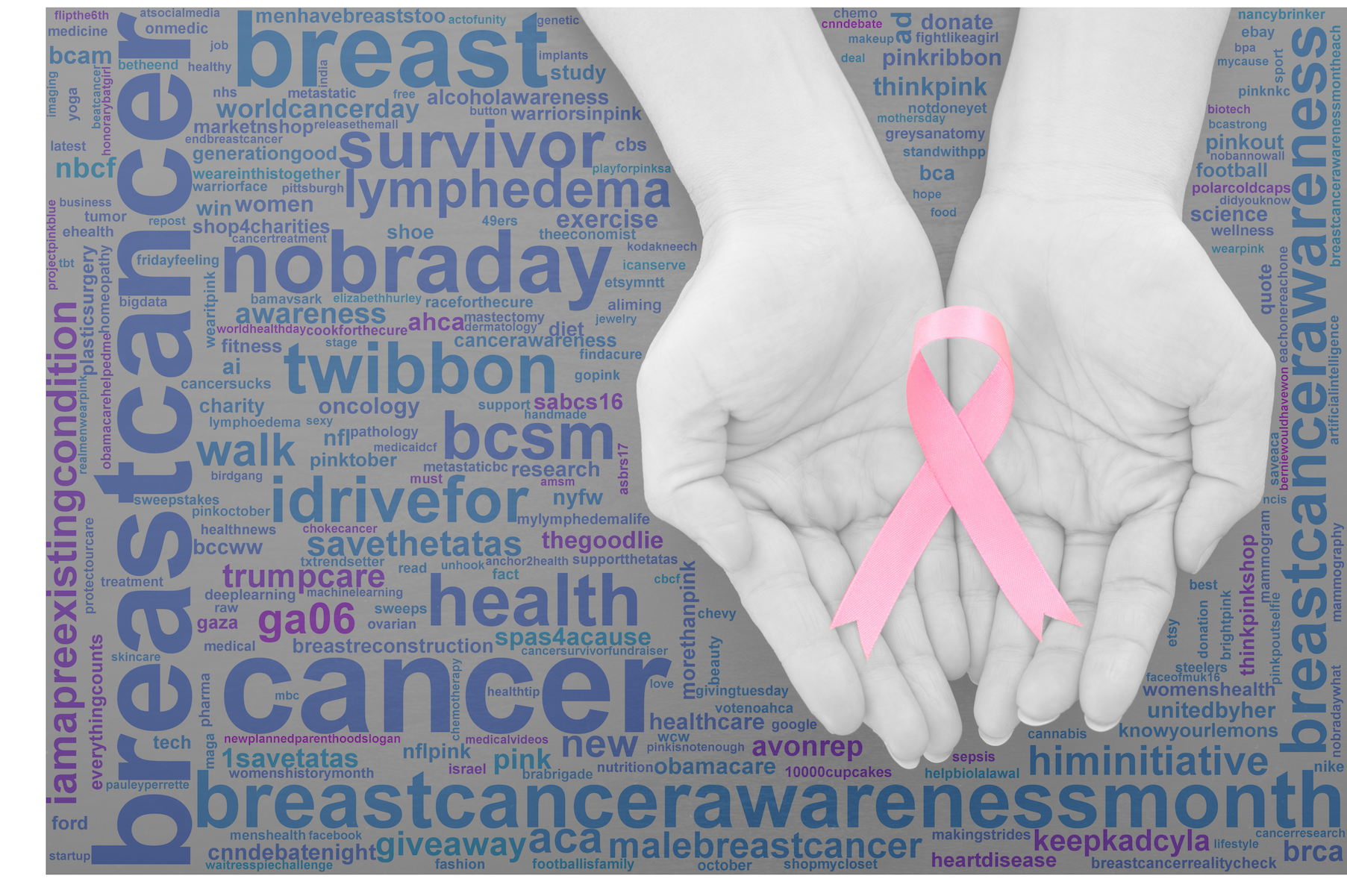}
\caption*{\textbf{Figure A4:} This word cloud displays the most prominent hashtags from all collected ``breast cancer'' tweets.  The hashtag sizes are proportionate to their relative frequencies and colors represent their average ambient happiness scores.  Here, light blue terms appear with the most positive LabMT words while purple hashtags appear with relatively more negative terms.}
\label{fig:HappCloud}
 \end{figure}

\end{document}